\documentclass[12pt,a4paper]{article}
\usepackage{amsmath,amssymb,mathrsfs,framed,esint,slashed}
\usepackage{natbib}
\usepackage{wasysym}
\usepackage{graphicx}
\usepackage{color}
\usepackage[all]{xy}
\usepackage{authblk}
\newtheorem{theorem}{Theorem}[section]

\newtheorem{lemma}[theorem]{Lemma}
\newtheorem{remark}[theorem]{Remark}

\newcommand{\pp}[2]{\frac{\partial #1}{\partial #2}}

\newcommand{\rem}[1]{}

\newcommand{\de}{{\rm d}}

\newcommand{\bq}{{\boldsymbol{q}}}
\newcommand{\bv}{{\boldsymbol{v}}}
\newcommand{\bp}{{\boldsymbol{p}}}

\newcommand{\bX}{{\mathbf{X}}}

\newcommand{\bu}{{\boldsymbol{u}}}
\newcommand{\bxi}{{\boldsymbol{\xi}}}

\newcommand{\bmu}{\boldsymbol{\mu}}
\newcommand{\bOm}{\boldsymbol{\Omega}}
\newcommand{\beq}{\begin{equation}}
\newcommand{\eeq}{\end{equation}}
\newcommand{\bal}{\begin{align}}
\newcommand{\eal}{\end{align}}
\newcommand{\mso}{\mathfrak{so}}

\newcommand{\todo}[1]{\vspace{5 mm}\par \noindent
\framebox{\begin{minipage}[c]{0.95 \textwidth}
\tt #1 \end{minipage}}\vspace{5 mm}\par}

\begin{document}

\title{Variational Neural Networks for Observable Thermodynamics (V-NOTS)}
\rem{ 
\author[1]{Christopher Eldred}
\email{celdred@sandia.gov}
\author[2]{Fran\c{c}ois Gay-Balmaz}
\email{francois.gb@ntu.edu.sg}
\author[3]{Vakhtang Putkaradze\corref{cor1}}
\cortext[cor1]{Corresponding author}
\ead{putkarad@ualberta.ca}

\affil[1]{organization = {Computer Science Research Institute}, 
addressline={Sandia National Laboratory, 1450 Innovation Pkwy SE}, city={Albuquerque}, state = {NM}, postcode={87123}, country={USA}}

\affil[2]{organization = {Division of Mathematical Sciences}, 
addressline={Nanyang Technological University}, city={637371}, country={Singapore}}

\affil[3]{organization={Department of Mathematical and Statistical Sciences},
            addressline={University of Alberta}, 
            city={Edmonton},
            postcode={T6G 2G1}, 
            state={Alberta},
            country={Canada}}
            }
            
\date{\today}
\author[1]{Christopher Eldred}
\author[2]{Fran\c{c}ois Gay-Balmaz}
\author[3]{Vakhtang Putkaradze \thanks{Corresponding author} }

\affil[1]{Computer Science Research Institute, 
Sandia National Laboratory, 1450 Innovation Pkwy SE,  Albuquerque, NM,  87123,  USA, email: celdred@sandia.gov}

\affil[2]{ Division of Mathematical Sciences, 
Nanyang Technological University,  637371, Singapore, email: francois.gb@ntu.edu.sg}

\affil[3]{ Department of Mathematics, 
University of Alabama, Tuscaloosa, AL 35401
USA, email: vputkaradze@ua.edu
}
\maketitle
\begin{abstract}
    Much attention has recently been devoted to data-based computing of evolution of physical systems. In such approaches, information about data points from past trajectories in phase space is used to reconstruct the equations of motion and to predict future solutions that have not been observed before. However, in many cases, the available data does not correspond to the variables that define the system's phase space. We focus our attention on the important example of dissipative dynamical systems. In that case, the phase space consists of coordinates, momenta and entropies; however, the momenta and entropies cannot, in general, be observed directly. To address this difficulty, we develop an efficient data-based computing framework based exclusively on observable variables, by constructing a novel approach based on the \emph{thermodynamic Lagrangian}, and constructing neural networks that respect the thermodynamics and guarantees the non-decreasing entropy evolution.  We show that our network can provide an efficient description of phase space evolution based on a limited number of data points and a relatively small number of parameters in the system. 
    \\
    \textbf{Keywords:} Dissipative systems, variational methods, structure-preserving machine learning, thermodynamics systems. 
\end{abstract}

\rem{ 
\section*{Statement of significance}
Predicting the motion of a physical system based on the data about its motion, and limiting information about the systems and forces acting on it, is a challenging and important problem. The problem becomes even more difficult when we cannot directly observe the phase space variables (e.g., momenta and entropies) and instead have to use different variables we can observe (correspondingly, velocities and temperatures). We show that, surprisingly, this challenging problem can be solved for a general class of systems using the Variational Neural Networks for Observable Thermodynamics (V-NOTS). We show that V-NOTS preserve the variational structure of the problem, are thermodynamically consistent, and can solve quite challenging problems with only partial information about the forces and the system. 
} 
\section{Introduction} 

\subsection{Background: general approaches to machine learning of systems with structure }

Many recent studies have been dedicated to the data-based discovery of differential equations that describe a physical system and predict its evolution. Incorporating physical insights into the model description can improve accuracy and reduce the number of data points needed to achieve a desired level of precision. In general, the problem of equation discovery is formulated in the following way.

\paragraph{On equation discovery through data: a continuous approach.}
Suppose that we know the phase space of the system and have one or several sequences of data obtained from experimental measurements or computer simulation of the system, $\bu_i = \bu(t_i)$, at time points $(t_1, \ldots, t_N)$. The goal is to determine the equations of motion 
$\bu'=\boldsymbol{f}(\bu)$ governing the system. 

In this approach, the differential equations are represented by a neural network, whose flow field approximates, in some optimal sense, the available data \cite{schaeffer2017learning,raissi2018deep,rackauckas2020universal}. While the general approach to equation discovery is generally well understood, the case in which the equations possess a specific structure remains challenging and requires more precise insight into the system. 

In this paper, we focus on the case where the equations describing the system are known to arise from a mechanical-thermodynamical system experiencing friction. To incorporate this type of knowledge, the structure of the equations must be included in the learning procedure,  which guarantees preservation of the dissipative nature of the system. The simplest way to write this system in the form $\bu'=\boldsymbol{f}(\bu)$ is to use the generalized coordinate-momentum representation with $\bu = (\bq, \bp)$, 
which describe the mechanical aspects of the system. In addition to these mechanical variables, purely thermodynamic variables such as entropy, mole numbers, and volumes must also be introduced \cite{StSc1974}. In the particular case of a simple system, only a single entropy variable $S$ is required in addition to the  variables $(\bq,\bp)$. For such systems, given a Hamiltonian function 
$H(\bq,\bp,S)$ and a friction force 
$\boldsymbol{F}^{\rm fr}(\bq, \dot{\bq}, S)$, the evolution can be expressed in the form:
\begin{equation} 
\dot{\bp} = - \pp{H}{\bq} + \boldsymbol{F}^{\rm{fr}} \, , \quad \dot \bq = \pp{H}{\bp} \, , \quad  T \dot S = - \pp{H}{\bp} \cdot \boldsymbol{F}^{\rm{fr}}  \, , \quad T := \pp{H}{S}\,  
\label{Dissipative_system_general_pq}. 
\end{equation} 
Here, the gradients of the Hamiltonian give the frictionless part of the system evolution, and the friction force must be chosen so that the total entropy (or, in the case of \eqref{Dissipative_system_general_pq}, the single entropy $S$) is non-decreasing over time. 
The dot always denotes differentiation with respect to time. The function $\boldsymbol{F}^{\rm{fr}}$ represents all non-conservative forces, while all potential forces are assumed to be incorporated into the Hamiltonian $H$ through the potential energy. We assume that the Hamiltonian function, which represents the total energy of the system, has no explicit time dependence and that the system is isolated.  

From \eqref{Dissipative_system_general_pq} we see that the total energy $H$ is conserved  for any friction force $\boldsymbol{F}^{\rm{fr}}$, i.e., $\dot H =0$. From the second law of thermodynamics, the force $\boldsymbol{F}^{\rm{fr}}$ must be dissipative, so that $\dot S \geq 0$. Any numerical scheme or machine-learning approximation to the solution that rigorously enforces $\dot S \geq 0$ is called thermodynamically consistent.  Neither the Hamiltonian nor the friction function is assumed to be known.

\paragraph{Metriplectic/GENERIC approach to describing dissipative systems.} A widely used framework for analyzing systems  of the form \eqref{Dissipative_system_general_pq} is the metriplectic formulation \cite{morrison1986paradigm,morrison2024}, also known as the GENERIC approach \cite{grmela1997dynamics,ottinger1997dynamics,ottinger2005beyond}. Although this framework differs from the approach developed in this paper, it is extensively used in the literature. For this reason, we briefly review it here in order to clarify the distinction.
In the GENERIC/metriplectic formalism, the dissipative system \eqref{Dissipative_system_general_pq} is written in the particular form  
\begin{equation}
\begin{aligned}
      \dot \bu & = \mathbb{L}(\bu) \pp{E}{\bu} + \mathbb{M}(\bu) \pp{S}{\bu} \, , \quad 
  \mbox{subject to } \quad  \mathbb{L}(\bu) \pp{S}{\bu} = \mathbb{M}(\bu) \pp{E}{\bu} = 0 \, ,
  \\ 
  \mathbb{L}(\bu) & = - \mathbb{L}(\bu){^T},\quad 
  \mathbb{M}(\bu)  \mbox{ symmetric positive semi-definite.}
\end{aligned} 
\label{}
\end{equation}
Here $E(\bu)$ denotes the energy and $S(\bu)$ the entropy. 
The matrices $\mathbb{L}(\bu)$ and $\mathbb{M}(\bu)$ define, respectively, antisymmetric and symmetric brackets on functions of $\bu$ as follows: 
\begin{equation}
\{ F, G \} = \pp{F}{\bu} \cdot \mathbb{L}(\bu) \pp{G}{\bu} \, , \quad 
( F, G ) =\pp{F}{\bu} \cdot \mathbb{M}(\bu) \pp{G}{\bu} .
    \label{Brackets_metriplectic}
\end{equation}
For genuine metriplectic or GENERIC systems, the bracket $\{F, G \}$ is required to be a Poisson bracket; in particular, it must satisfy the Jacobi identity. In practice, however, this requirement is not always enforced in the literature. 
From \eqref{Brackets_metriplectic}, it follows that the energy is conserved along solutions, while the entropy is non-decreasing in time: 
\begin{equation}
    \begin{aligned}
        \dot E & = \{ E, E \} + (S, E) = \pp{S}{\bu} \cdot \mathbb{M}(\bu) \pp{E}{\bu} =0 
        \\ 
        \dot S &  = \{ S, E \} + ( S, S) = (S,S) \geq 0 \, . 
    \end{aligned}
\end{equation}
Structure-preserving integrators for metriplectic brackets have been studied in considerable details; see, for example, \cite{ottinger2018generic,bloch2024metriplectic}. There has also been significant interest in incorporating these structures into machine learning approaches. 

\paragraph{Equations discovery for metriplectic systems.} 
Given a sequence of data $\left\{\bu_k\right\}_{k=1}^N$ sampled at time points $t_k = k \Delta t$, the goal is to identify the matrices $\mathbb{M}$ and $\mathbb{L}$, together with the functions $E$ and $S$. One of the earliest works on data-driven modeling of metriplectic systems is \cite{lee2021machine}, which assumed a particular learnable representation for the matrices $\mathbb{L}$ and $\mathbb{M}$ involving the gradients of $E$ and $S$, guaranteeing the antisymmetry of $\mathbb{L}$ and the positive definiteness of $\mathbb{M}$. This work was followed closely by the GFINNs paper \cite{zhang2022gfinns}, which considered the case where the unknown quantities depend on the phase-space variable $\bu$ with a prescribed structure. More recently, \cite{gruber2024efficiently} considered more general metriplectic matrices that depend on the gradients of the energy, ensuring that the matrix $\mathbb{M}$ has $\nabla E$ in its kernel. 
These works are consistent with recent theoretical advances in the understanding of metriplectic systems \cite{morrisonUTA}, and their generalization into metriplectic 4-brackets \cite{morrison2024}. 

\begin{figure}
    \centering
    \includegraphics[width=0.5\linewidth]{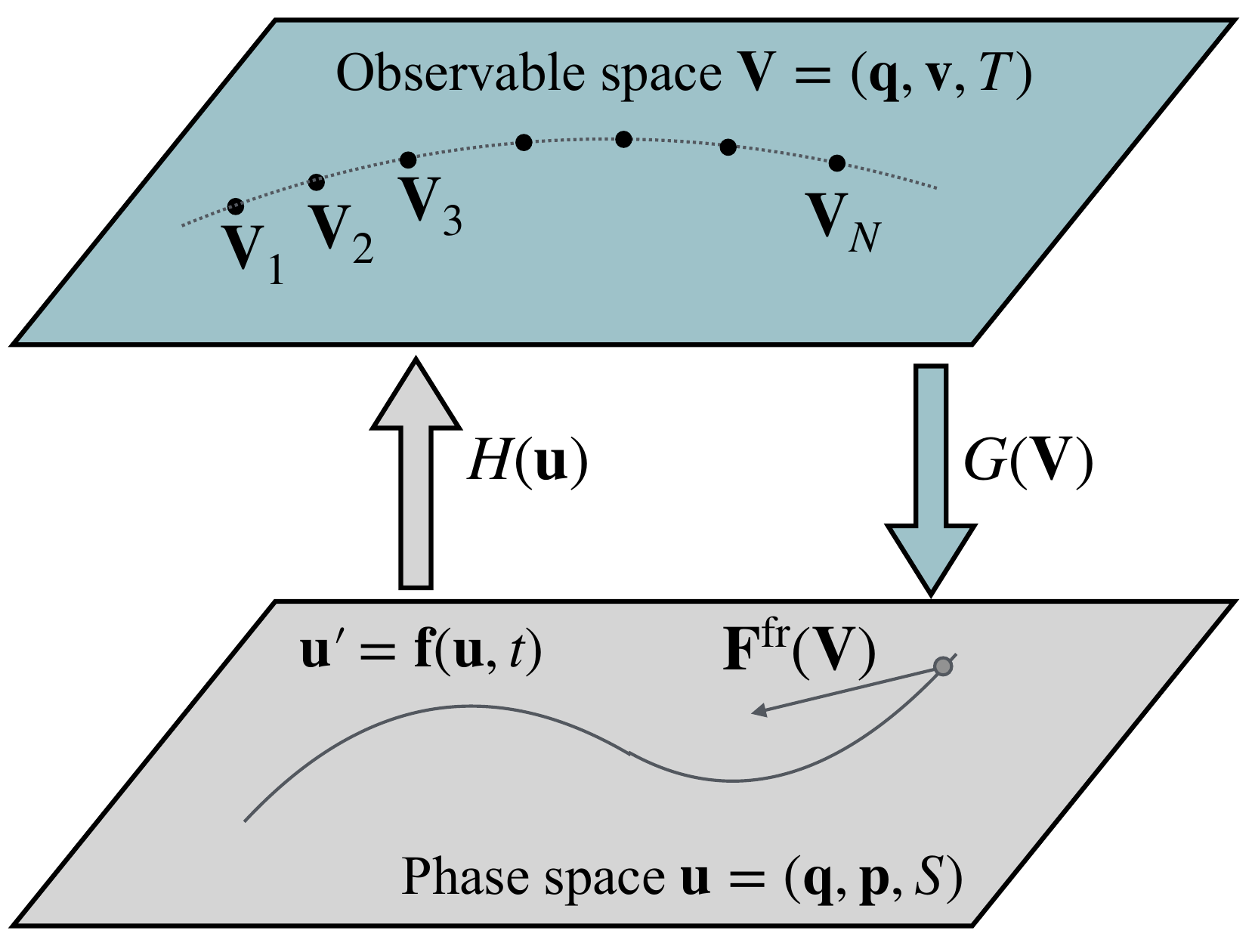}
    \caption{Illustration of the difference between evolution in observable and non-observable variables. The evolution equation \eqref{Dissipative_system_general_pq} is written in terms of the phase-space variables $\bu = (\bq,\bp,S)$, where the momentum $\bp$ and entropy $S$ are not directly observable in experiments without additional information. In contrast, the observable data consist of the coordinates $\bq$, velocities $\bv$, and temperatures $T$. The friction force is also dependent on the observable variables, although it is acting as a part of equation \eqref{Brackets_metriplectic}. The observable variables $(\bv,T)$ and the non-observable variables $(\bp,S)$ are related to each other by the derivatives of the Hamiltonian $H(\bq,\bp,S)$, which contains both mechanical and thermal contributions. In Section \ref{sec:observables} we introduce a thermal Lagrangian $G(\bq,\bv,T)$ which allows to connect the dynamics in observable variables $(\bq,\bv,T)$ with the phase space description in terms of $\bu$.}
    \label{fig:observable_nonobservable}
\end{figure}

It was already noted in \cite{lee2021machine} that entropy is an unobservable variable and therefore should not be included in the training dataset. We take this observation one step further by noting that momentum is also not directly observable and thus should not be included in the dataset either. This substantially complicates the application of metriplectic approach to practical systems.
The distinction between observable and unobservable variables, the resulting difficulties in predicting the system evolution, and the solution proposed in this paper, which is distinct from the metriplectic/GENERIC approach, are illustrated in Figure~\ref{fig:observable_nonobservable}.

\paragraph{Observable and non-observable data.} Dissipative systems present an additional challenge that complicates the application of data-based metriplectic methods. The difficulty arises from the nature of the observable data. In the formulation \eqref{Brackets_metriplectic}, it is typically assumed that the dataset $(\bu_1,\ldots,\bu_N)$ contain sufficient information about all (or nearly all) phase-space variables $\bu$ at given times $t_i$. For mechanical systems, however, this assumption fails in an essential way. A dissipative mechanical system is formulated in terms of coordinates, momenta, and entropy, which together form the phase-space variable $\bu$. While the coordinates can be directly observed in experiments, the momenta generally cannot. Even for a single particle, observations of positions and velocities contain no information about the particle's mass and therefore do not determine its momentum. Any conclusion on the mass of the particle is only possible by either the knowledge of the forces acting on the particle, or the form of mechanical energy. Similarly, entropy cannot be inferred from direct observations of the system without knowledge of the form of the thermal energy. If the explicit expression of the total energy (the Hamiltonian) is known, the observable velocities and temperatures are related to the partial derivatives of the Hamiltonian with respect to the unobservable momenta and entropy, as illustrated in the example given in equation \eqref{Dissipative_system_general_pq} below. In practice, however, the Hamiltonian is typically unknown. Consequently, \emph{the observable variables correspond to partial derivatives of an unknown function with respect to unobservable variables}. Moreover, as discussed below, friction forces are formulated in terms of observable variables rather than phase-space variables. This lack of direct observability of the phase-space variables presents a fundamental challenge for data-based computing of dissipative systems. Addressing this difficulty is the primary goal of this paper.

\paragraph{Discovering equations vs mappings in phase space: continuous and discrete approaches.} It is useful to distinguish between two different approaches to data-based computing.

The works cited above on data-based approximation of general system \eqref{Dissipative_system_general_pq} focus on computing the equations of motion, or, in formal terms, the system's infinitesimal flow. 
In the context of systems with structure, such as canonical Poisson systems, a substantial body of literature has also developed methods for computing equations of motion that are rigorously Hamiltonian \cite{greydanus2019hamiltonian}. Once such equations are derived, typically in the form of a neural network approximation, the resulting system must still be solved numerically, for example, using a structure-preserving integrator \cite{marsden2001discrete}.

In contrast, other works have developed a direct approximation of the mapping of phase space over a fixed time interval $\Delta t$. This approach exploits the structure of phase-space mappings associated with canonical Hamiltonian systems \cite{jin2020sympnets} and more general Poisson systems \cite{jin2022learning,vaquero2023symmetry,vaquero2024designing,eldred2024lie,eldred2025clpnets}. One could call that approach \emph{discrete} data-based computing. Learning the phase-space mapping has the advantage that  it allows a much more direct computation of solutions, without the need to compute the solution using a 'continuous' solver. However, compared with Hamiltonian systems and their generalizations, very little is known about the structure of phase-space mappings generated by dissipative systems. Thus, there has been little progress in the discrete approach to the dissipative systems, compared to the case of Hamiltonian systems.

\paragraph{Differences from the metriplectic/GENERIC approach.} 
There are several fundamental differences between the method proposed in this paper and the metriplectic/GENERIC approach. 
\begin{enumerate} 
\item Our approach uses variational schemes for machine learning based on structure-preserving methods derived from the Hamilton-like critical action principle \eqref{VCond}-\eqref{PC}, rather than relying on a bracket formulation. 
\item It is based on variational integrators stemming from the Hamilton-like action principle, rather than on integrators derived from bracket structures. 
\item Our formulation is expressed in terms of \emph{observable} variables (coordinates, velocities and temperatures), rather than phase space variables used in the metriplectic/GENERIC approach. 
\end{enumerate}
The latter distinction will be of particular importance. Recent work has shown that the variational thermodynamics framework that forms the foundation of our method (described in Section~\ref{sec:var_thermo}) yields, formally, equations of motion equivalent to those obtained in 
metriplectic formulations of nonequilibrium thermodynamics \cite{gay2020variational,ElGB2020,carlier2024metriplectic}. An important difference, however, is that 
the metriplectic approach for data-based computing must necessarily assume that the values of the phase space variables $\bu$ are known \cite{zhang2022gfinns,gruber2024efficiently,gruber2024reversible}. These variables necessarily involve momenta and entropies, which, as noted above, are not directly observable in physical experiments. 
Thus, it is difficult to see how data-based computing of thermodynamic systems based on the GENERIC/metriplectic framework can be carried out when only observable quantities are available. The goal of this paper is therefore to develop an alternative data-based approach for dissipative systems, grounded on Lagrangian functions and on the variational formulation of thermodynamics. 
The key advantage of the variational thermodynamics framework is that it allows us to: 
\begin{enumerate}
\item Rewrite the variational formulation entirely in terms of the observable variables corresponding to the thermodynamics Lagrangian representation (position, velocities, temperatures).
    \item Use the generalized action principle to design variational integrators that can be incorporated in machine-learning algorithms.  
\end{enumerate}

\paragraph{Discussion on variational approaches for mechanical systems with friction.} Before presenting our approach, it is useful to clarify what is meant by a \emph{variational} approach, as the term is used in different ways in the literature. 
\\ 
\emph{Variational in the sense of Onsager.} In 1931, L. Onsager published two fundamental papers \cite{onsager1931reciprocal1,onsager1931reciprocal2}, introducing what is now known as the \emph{Onsager variational principle}. That approach computes the \emph{instantaneous} dissipation due to all processes, and rate of change of the internal energy and dissipative forces, and takes the minimum of that functional \emph{at a given time} with respect to the generalized velocities. Onsager's principle is local in time and does not involve an integral over time, in contrast to action-based variational formulations. It is primarily concerned with dissipative processes near equilibrium, emphasizing entropy production and irreversible effects. Consequently, the evolution equations derived from Onsager's principle do not, in general, conserve mechanical energy. Variational numerical schemes based on this idea have been studied for a long time, for example through time-discretization of energy functionals, such as the De Giorgi schemes \cite{deGiorgi1993new}, see also \cite{chen2025onsager} for further developments. Subsequent work utilized these principles for machine learning, constructing structure-preserving numerical schemes  such as \emph{Variational Onsager Neural Networks} (VONNs) \cite{huang2022variational} and their extensions \cite{qiu2025bridging}. 
Related studies  \cite{sharma2024preserving,sharma2024lagrangian} have developed non-intrusive methods to learn Lagrangian mechanics from data, where mechanical energy is dissipated through friction forces that are assumed to come from a Rayleigh dissipation function.  

\paragraph{Difference from Onsager-based approaches.} Our work differs from Onsager-based approaches in two key respects:
\begin{enumerate} 
\item Unlike Onsager-based works, we explicitly include the evolution of the thermal component $S$. This allows the system to conserve total energy through conversion between internal and mechanical energy, whereas in Onsager-based approaches only the dissipation of mechanical energy is captured.
\item As we state below, our variational approach described in Section~\ref{sec:var_thermo} is based on a Hamilton-like critical action principle, in which the functional to be extremized involves an integral over time.
\end{enumerate}


\paragraph{Problem statement: data-based computing of dissipative systems from observable data.} This paper solves the following problem. Suppose we observe a dissipative thermodynamic system, and we record $N$ pairs of observable data $\big\{ (\bX_k^0,\bX_k^f) \big\}_{k=1}^N$ corresponding to the beginning and end of a time interval $\Delta t_k$. The observable data $\bX$ do not correspond directly to the phase space variables $\bu=(\bq,\bp,S)$ in \eqref{Dissipative_system_general_pq}. Our goal is to reconstruct the dynamics in the whole space while respecting the laws of thermodynamics.

\paragraph{Novelty of this paper.} This paper presents the following novel results: 
\begin{enumerate}
\setlength{\itemsep}{0pt}
    \item We derive a novel formulation of dissipative systems in terms of the thermal analogue of Legendre transform of the Hamiltonian, which we call the \emph{thermal Lagrangian}.
    \item We develop a strictly dissipative neural network architecture for modeling dissipative forces. This architecture is capable of representing all dissipative forces and guarantees strict dissipation for all parameter values.
    \item We derive a discrete approach to data-based simulations of dissipative systems using the method of variational integrators for dissipative systems, preserving the dissipative structure of the phase flow. 
    \item We show that for particular problems, data-based computing based solely on observable variables is not solvable as it does not contain enough information. However, once sufficient information about the system is provided, our method yields accurate long-term approximation of the dissipative dynamics. 
\end{enumerate}

\paragraph{A short digression on systems with symmetry and observable variables.}
When the configuration manifold of the system is a Lie group $\bq = \boldsymbol{g} \in G$, and the system is invariant with respect to that group, the evolution of the system can be written in terms of the reduced momentum $\boldsymbol{\mu} = \boldsymbol{g}^{-1} \bp $ or $\boldsymbol{\mu} = \bp \boldsymbol{g}^{-1}$ depending on the type of symmetry (left or right). The momentum $\boldsymbol{\mu}$ belongs to $\mathfrak{g}^*$,  the dual of the Lie algebra $\mathfrak{g}$ of $G$. Similarly, the friction force $\boldsymbol{F}^{\rm fr}$ may be invariant, giving rise to its reduced expression $\boldsymbol{f}^{\rm fr}\in \mathfrak{g}^*$. In that case the Hamiltonian only depends on $\boldsymbol{\mu}$ and $S$ so we can write $H(\boldsymbol{g},\boldsymbol{p},S) = h(\boldsymbol{\mu}, S)$. Consequently, the system \eqref{Dissipative_system_general_pq} can be written in terms of $\boldsymbol{\mu}$ and $S$ only as 
\begin{equation} 
\dot {\boldsymbol{\mu}} \mp  {\rm ad}^*_{\pp{h}{\boldsymbol{\mu}}} \boldsymbol{\mu}= \boldsymbol{f}^{\rm{fr}}  \, , \quad T \dot S = -   \pp{h}{\boldsymbol{\mu}}\cdot  \boldsymbol{f}^{\rm fr} , \quad T:= \pp{h}{S},
\label{mu_eq_gen} 
\end{equation} 
see \cite{CoGB2020}.
In \eqref{mu_eq_gen}, one chooses the minus sign for left-invariant systems and the plus sign for right-invariant systems. We will develop the theory of machine learning for \eqref{mu_eq_gen} in parallel with the canonical system \eqref{Brackets_metriplectic}. 

The variable $\boldsymbol{\xi}=\pp{h}{\boldsymbol{\mu}}\in \mathfrak{g}$ is the reduced (e.g. body or spatial) velocity, and the concept of temperature is unchanged. 

\paragraph{On the expression for the friction force.} One often approximates the friction force, for example using Stokes' law, where the force is proportional to velocity. In reality, however, the friction may not allow for a closed algebraic expression for all velocities and temperatures. We further assume that the expression for the friction force is unique as a function of the observable variables. In reality, there may be multiple expressions for the friction force, \emph{e.g.}, when a body is in an external fluid flow that transitions from laminar to recirculating flow. We do not consider these cases, as the formal non-uniqueness of the friction is due to lack of sufficient details in the consideration of the system containing the body and the fluid. In this paper, we assume that all essential components in the system are taken into account by equations \eqref{Dissipative_system_general_pq} or their symmetry-reduced analogues \eqref{mu_eq_gen}.

\color{black}

\section{Background: Variational approach to dissipative systems and variational integrators} 
\label{sec:var_thermo}
In this section, we assume that both the functional forms and the derivatives of the Hamiltonian 
$H$ and $\boldsymbol{F}^{\rm fr}$ in \eqref{Dissipative_system_general_pq}, or $h$ and $\boldsymbol{f}^{\rm fr}$ in \eqref{mu_eq_gen}, are explicitly known in terms of the independent variables. 
When we apply these ideas to practical problems in the data-based computing in Section \ref{sec:physics_dissipative_NN}, we will compute the Hamiltonians and friction forces using a neural network, while the derivatives of the Hamiltonian will be obtained via automatic differentiation. 

\subsection{Variational approach to thermodynamics} 

We present a short introduction of the variational theory of thermodynamics developed in \cite{GBYo2017a,gay2020variational,GBYo2022,GBYo2023}. In that approach, the equations of motion of thermodynamic systems can be derived by a variational principle which extends the Hamilton principle of mechanics.

\paragraph{Simple thermodynamic systems.} For systems with a single entropy, see \eqref{Dissipative_system_general_pq}, given the Lagrangian $L(\bq, \dot{\bq},S)$ and the friction force $\boldsymbol{F}^{\rm fr}(\bq, \dot{\bq}, S)$, one considers the variational principle
\begin{equation}\label{VCond}
\delta \int_0^T L( \bq, \dot{\bq},S) {\rm d}t=0,
\end{equation}
where the critical curve, resp., the variations $\delta\bq$, $\delta S$, satisfy the constraint
\begin{equation}\label{PC}
\frac{\partial L}{\partial S}\dot S = \boldsymbol{F}^{\rm fr}\cdot \dot\bq ,\quad\text{resp.},\quad
\frac{\partial L}{\partial S}\delta S = \boldsymbol{F}^{\rm fr}\cdot \delta\bq
\end{equation}
with $\delta\bq(0)=\delta\bq(T)=0$. This procedure results in equations
\[
\frac{d}{dt}\frac{\partial L}{\partial\dot{\bq}}- \frac{\partial L}{\partial\bq}= \boldsymbol{F}^{\rm fr}, \qquad \frac{\partial L}{\partial S}\dot S = \boldsymbol{F}^{\rm fr}\cdot \dot\bq ,
\]
which are readily seen to be equivalent to \eqref{Dissipative_system_general_pq} with $H$ the Hamiltonian associated to $L$, assumed to be nondegenerate. One can also obtain \eqref{Dissipative_system_general_pq} directly, by considering the variational principle
\[
\delta \int_0^T \big(\bp \cdot \dot{\bq} - H( \bq, \bp,S) \big){\rm d}t=0\quad\text{subject to}
\quad -\frac{\partial H}{\partial S}\dot S = \boldsymbol{F}^{\rm fr}\cdot \dot\bq , \;\;
-\frac{\partial H}{\partial S}\delta S = \boldsymbol{F}^{\rm fr}\cdot \delta\bq,
\]
see \cite{GBYo2022}. Note that this variational principle is of \textit{d'Alembert type}, as used for nonholonomic mechanical systems, in the sense that the variations are constrained, and such constraints are naturally inherited by the constraints imposed on the critical curve. The major difference with the original d'Alembert principle is that in the thermodynamics applications, the constraints are nonlinear in velocities. The variational theory readily extends to the case of several entropies, which we will need for the particular case of a double piston considered in Section~\ref{sec:adiabatic-piston}. A brief presentation of variational thermodynamics for multi-entropy case can be found in Appendix~\ref{app:mutl_entropies}. 

\paragraph{Invariant thermodynamical systems on Lie groups.} We assume that the configuration manifold is a Lie group and the Lagrangian is left-invariant: $L(\boldsymbol{g}, \dot{\boldsymbol{g}}, S)= L(\boldsymbol{h}\boldsymbol{g}, \boldsymbol{h}\dot{\boldsymbol{g}}, S)$, for all $\boldsymbol{h}\in G$, so that it can be expressed in terms of the reduced velocity $\boldsymbol{\Omega}=\boldsymbol{g}^{-1}\dot{\boldsymbol{g}}$ as $L(\boldsymbol{g}, \dot{\boldsymbol{g}}, S)=\ell(\boldsymbol{\Omega},S)$. The equations of motion follow as before by applying \eqref{VCond}-\eqref{PC}. From invariance, however, we can reformulate the variational principle directly in terms of the reduced Lagrangian $\ell(\boldsymbol{\Omega},S)$. One gets
\begin{equation}\label{VCond_EP}
\delta \int_0^T \ell( \boldsymbol{\Omega},S) {\rm d}t=0
\end{equation}
subject to the following constraints on the curves, resp., on the variations
\begin{equation}\label{PC_VC_EP}
\frac{\partial \ell}{\partial S}\dot S = \boldsymbol{f}^{\rm fr}\cdot \boldsymbol{\Omega}, \quad\text{resp.}\quad \frac{\partial \ell}{\partial S}\delta S = \boldsymbol{f}^{\rm fr}\cdot \boldsymbol{\Sigma}, \qquad \delta \boldsymbol{\Omega}= \dot{\boldsymbol{\Sigma}}+[\boldsymbol{\Omega},\boldsymbol{\Sigma}],
\end{equation}
where $\boldsymbol{\Sigma}$ is an arbitrary curve in $\mathfrak{g}$ vanishing at $t=0,T$, see \cite{CoGB2020}. This results in the equations
\[
\frac{d}{dt}\frac{\partial \ell}{\partial\boldsymbol{\Omega}}= \operatorname{ad}^*_{\boldsymbol{\Omega}}\frac{\partial \ell}{\partial\boldsymbol{\Omega}}+ \boldsymbol{f}^{\rm fr}, \qquad \frac{\partial \ell}{\partial S}\dot S = \boldsymbol{f}^{\rm fr}\cdot \boldsymbol{\Omega}. 
\]
These equations recover \eqref{mu_eq_gen} when written in terms of the Hamiltonian $h(\boldsymbol{\mu},S)$ associated with $\ell(\boldsymbol{\Omega},S)$. Here $[\boldsymbol{\Omega},\boldsymbol{\Sigma}]$ denotes the Lie bracket on $\mathfrak{g}$, $``\cdot"$ denotes the pairing between $\mathfrak{g}$ and it dual, and $ \operatorname{ad}^*_{\boldsymbol{\Omega}}\boldsymbol{\mu}\cdot \boldsymbol{\Sigma}= \boldsymbol{\mu}\cdot [\boldsymbol{\Omega},\boldsymbol{\Sigma}]$ is the coadjoint operator.

In Section \ref{sec:physics_dissipative_NN} we will consider the question of the arguments of the dissipative force in more detail from the general principles of physics. 

\subsection{Variational integrators for thermodynamics}

To build the discrete flow underlying our method, we shall use the idea of variational integrators. For such methods, \cite{marsden2001discrete}, one first discretizes the Lagrangian and then computes the critical point of the associated discrete action functional, thereby giving the discretized equations. For a mechanical system with Lagrangian $L(\bq, \dot \bq)$, one considers a discrete Lagrangian $L_d(\bq_k, \bq_{k+1})\simeq \int_{t_k}^{t_{k+1}}L(\bq, \dot \bq){\rm d}t$. The discrete Hamilton principle $\delta \sum_{k=0}^{N-1} L_d(\bq_k, \bq_{k+1}) =0$, for variations $\delta\bq$ with $\delta\bq_0=\delta\bq_N=0$, then yields
\[
\partial_1 L_d(\bq_k,\bq_{k+1}) + \partial_2 L_d(\bq_{k-1},\bq_k)=0
\]
which, under suitability regularity conditions of $L_d$, defines the discrete flow.

As shown in \cite{gay2018variational}, this approach can be extended to the variational formulation of thermodynamics given in \eqref{VCond}-\eqref{PC}. Given $L(\bq, \dot\bq, S)$, $\boldsymbol{F}^{\rm fr}(\bq, \dot{\bq},S)$ and a finite difference map, $ (\bq_k,\bq_{k+1},S_k, S_{k+1})\mapsto \varphi (\bq_k,\bq_{k+1},S_k, S_{k+1})\simeq(\bq,\dot{q},S, \dot S)$, construct the associated discrete Lagrangian $L_q(\bq_k, \bq_{k+1}, S_k, S_{k+1})$, discrete friction forces, and discrete version of the constraint \eqref{PC}. The discrete variational principle is then $\delta \sum_{k=0}^{N-1}L_d(\bq_k, \bq_{k+1}, S_k, S_{k+1})=0$, with respect to variations $\delta\bq_k$ and $\delta S_k$ subject to the natural counterpart of \eqref{PC}, see \cite{gay2018variational} for details.

\paragraph{Variational integrator for simple thermodynamic systems.} By choosing the finite difference map $\varphi( \bq_k, \bq_{k+1}, S_k, S_{k+1})= ( \bq_k, S_k, (\bq_{k+1}-\bq_k)/h, (S_{k+1}-S_k)/h)$, the variational integrator gives the scheme
\begin{equation}
\left\{ 
\begin{array}{l}
\vspace{0.2cm}  \displaystyle\frac{1}{h}\frac{\partial L}{\partial\bv}(\bq_k, \bv_k, S_k) - \frac{1}{h}\frac{\partial L}{\partial\bv}(\bq_{k-1}, \bv_{k-1}, S_{k-1})-\frac{\partial L}{\partial \bq}(\bq_k, \bv_{k}, S_k)= \boldsymbol{F}^{\rm fr} \left(\bq _k, \boldsymbol{v}_{k}, S_k \right)\\
\displaystyle \frac{\partial L}{\partial S}(\bq _k , \bv_k, S_k ) \frac{S_{k+1}-S_k}{h}=  \boldsymbol{F}^{\rm fr} \left(\bq _k,\boldsymbol{v}_{k}, S_k \right) \cdot \boldsymbol{v}_{k},\qquad \boldsymbol{v}_k:= \frac{\bq _{k+1}-\bq_k}{h}.
\end{array}
\right.
\label{var_int}
\end{equation} 
Assuming the Lagrangian is nondegenerate with respect to the mechanical variables, it can be equivalently written using the Hamiltonian as
\begin{equation}
\left\{ 
\begin{array}{l}
\vspace{0.2cm}  \displaystyle\frac{\bq_{k+1}- \bq_k}{h} = \boldsymbol{v}_{k},\\
\vspace{0.2cm}  \displaystyle\frac{\bp_{k+1}-\bp_k}{h}=-\frac{\partial H}{\partial \bq}(\bq_k, \bp_{k+1}, S_k)+ \boldsymbol{F}^{\rm fr} \left(\bq _k, \boldsymbol{v}_{k}, S_k \right),\\
\displaystyle\frac{\partial H}{\partial S}(\bq _k , \bp_{k+1}, S_k ) \frac{S_{k+1}-S_k}{h}= - \boldsymbol{F}^{\rm fr} \left(\bq _k,\boldsymbol{v}_{k}, S_k \right) \cdot \boldsymbol{v}_{k},\qquad  
\boldsymbol{v}_k:= \displaystyle\pp{H}{\bp}\left(\bq _{k},\bp _{k+1}, S_k \right). 
\end{array}
\right.
\label{var_int_canonical}
\end{equation} 
In practice, we can rewrite the friction forces in terms of the temperature $T_k=T_k(\bq_k, S_k)$ instead of the entropy, using the expression of the internal energy.

\paragraph{Thermodynamic systems with several entropies.} In absence of heat exchanges, the same discrete variational approach as above gives the scheme 
\begin{equation}
\left\{ 
\begin{array}{l}
\vspace{0.2cm}  \displaystyle\frac{\bq_{k+1}- \bq_k}{h} = \boldsymbol{v}_{k},\qquad \frac{\bp_{k+1}-\bp_k}{h}=-\frac{\partial H}{\partial \bq} + \sum_{i=1}^P\boldsymbol{F}^{\rm fr (i)} ,\\
\displaystyle\frac{\partial H}{\partial S_i} \frac{(S_i)_{k+1}-(S_i)_k}{h}= - \boldsymbol{F}^{\rm fr(i) } \cdot \boldsymbol{v}_{k},\;\;\;\;i=1,...,P,\qquad\boldsymbol{v}_k:= \displaystyle\pp{H}{\bp},
\end{array}
\right.
\label{var_int_canonical_multiple}
\end{equation} 
where the variables of the partial derivatives of $H$ are $ (\bq _k , \bp_{k+1}, (S_1)_k,..., (S_P)_k )$ and those of the friction forces are $ (\bq _k , \bv_k, (S_1)_k,..., (S_P)_k )$.

\paragraph{Variational integrator for systems on Lie groups.} The variational discretization can be extended to thermodynamical systems on Lie groups \eqref{mu_eq_gen}, by selecting a local diffeomorphism $\tau: \mathfrak{g}\rightarrow G$ between the Lie group and its Lie algebra, such that $\tau (0)=e$, which approximates the exponential map. Then, under the left invariance assumption, the discrete Lagrangian can be written as $L_d(\mathbf{g}_k, \mathbf{g}_{k+1},S)= \ell_d( \boldsymbol{\Omega}_k, S_k)$ with $\boldsymbol{\Omega}_k= \frac{1}{h}\tau^{-1}( \mathbf{g}_k^{-1}\mathbf{g}_{k+1})$. The discrete variational principle discussed above can be expressed in terms of the reduced Lagrangian $\ell_d$ and reduced variable $\boldsymbol{\Omega}_k$, see \cite{CoGB2020}. For instance, for $G=SO(3)$, using for $\tau$ the Cayley approximation, writing $\Omega=\widehat{\boldsymbol{\Omega}}$ and $\mu=\widehat{\boldsymbol{\mu}}$ the skew symmetric $3\times 3$ matrices associated to the angular velocity and momentum 3-vectors $\boldsymbol{\Omega}$ and $\boldsymbol{\mu}$, this yields
\begin{equation}
\left\{ 
\begin{aligned}
 &  \frac{\bmu_{k+1} - \bmu_k}{h}  +  \frac{1}{2}\big([ \mu_{k+1}, \Omega_{k+1} ] + [ \mu_{k}, \Omega_{k}]\big)^\vee \\ 
 & \qquad \qquad - \frac{h}{4} \left( \Omega_{k+1} \mu_{k+1} \Omega_{k+1} -\Omega_{k} \mu_{k} \Omega_{k} \right)^\vee +  \boldsymbol{f}^{\rm fr}_{k+1}  =0 
 \\ 
 &  \frac{S_{k+1}-S_k}{h} = \frac{1}{T_k}   \bOm_k \cdot  \boldsymbol{f}_k  \, , \quad 
 \bOm_k:= \pp{h}{\bmu}(\bmu_k,S_k) \, , \quad  T_k = \pp{h}{S}(\bmu_k,S_k).
\end{aligned}
\right. 
    \label{var_integrator_reduced}
\end{equation}
Here, we have defined $a^\vee= \boldsymbol{a}$ to be the inverse hat map, which transforms antisymmetric $3 \times 3$ matrices to vectors in $\mathbb{R}^3$.

\section{Formulation of the problem in terms of observable data}
\label{sec:observables}
So far, we have formulated the equations of motion in terms of the quantities $(\bq, \bp, S)$. We assumed that, at discrete time points $t_0, t_1, \ldots t_N$, these quantities can be determined and used for learning. The coordinates $(\bq_1, \ldots, \bq_N)$ are certainly observable, but the momenta $(\bp_1, \ldots, \bp_N)$ and entropies are difficult to observe in practice without extra information. Even for a point particle, observation of velocity does not yield momentum, since the mass of the particle cannot be assumed to be known \emph{a priori}. Similarly, observation of the temperature of an object does not yield entropy, since determination of the entropy from temperature requires knowledge of the equation of state and potential energy. 

The observable variables for the thermodynamics system are the coordinates $\bq$, velocities $\bv$ and temperatures $T$. Given a Hamiltonian $H(\bq, \bp, S)$, the velocities and temperatures are related to the momenta and velocities as 
\begin{equation}
\bv(\bq, \bp, S) = \pp{H}{\bp} \, , \quad T(\bq, \bp, S) = \pp{H}{S} \, . 
    \label{vT_pS}
\end{equation}
We want to rewrite our method so that it only uses the observable variables $(\bq, \bv, T)$ as independent variables. This task is naturally achieved by taking the Legendre transform in the variables $(\bp, S)$ to the variables $(\bv, T)$. We thus define the function $G(\bq, \bv, T)$ which is the Lagrangian version of the Helmholtz free energy: 
\begin{equation}
    G(\bq, \bv, T) = \bp (\bq, \bv, T) \cdot \bv + S (\bq, \bv, T) T - H \left(  \bq,\bp (\bq, \bv, T) ,  S (\bq, \bv, T) \right) ,
    \label{Gibbs_Lagr}
\end{equation}
also considered in \cite{GBYo2019}. Here we break with the convention from the thermodynamics literature that Helmholtz free energy is denoted with F, since we already use that symbol for the friction force. We hope that no confusion arises from this somewhat unconventional notation. We verify that the function $G(\bq, \bv, T)$ satisfies
\begin{equation}
    \bp = \pp{G}{\bv} \, , \quad S = \pp{G}{T} \, . 
    \label{G_deriv_pS}
\end{equation}
We now proceed to the variational principle which we formulate as 
\begin{equation}
\begin{aligned}
      & \delta \int_0^T ( G - T S) \mbox{d} t =0 \quad \mbox{subject to }
      \\
     & T \dot{S} = - \boldsymbol{F}^{\rm fr} \cdot \dot{\bq} \\ 
     &T \delta S = - \boldsymbol{F}^{\rm fr} \cdot \delta \bq  \,      
\end{aligned}
\label{Lagr_var_principle_thermo}
    \end{equation}
    The variational principle \eqref{Lagr_var_principle_thermo} gives: 
    \begin{equation}
    \begin{aligned}
        \int _0^T\left( - \frac{d}{d t}  \pp{G}{\bv} + \pp{G}{\bq} + \boldsymbol{F}^{\rm fr} \right) 
        \cdot \delta \bq + \left( \pp{G}{T}  - S \right) \delta T \de t = 0 \, .
    \end{aligned}
\label{deriv_variations}
    \end{equation}
Using \eqref{G_deriv_pS}, we notice that the term proportional to $\delta T$ vanishes under the integral and we get the thermodynamic analogue of Euler-Lagrange equations expressed in the observable variables $(\bq, \bv, T)$: 
    \begin{equation}
    \begin{aligned}
        \frac{d}{d t} &  \pp{G}{\bv} = \pp{G}{\bq} + \boldsymbol{F}^{\rm fr} \, , \quad \bv = \dot \bq 
        \\    
        T\frac{d}{d t} &  \pp{G}{T} = - \boldsymbol{F}^{\rm fr} \cdot \bv   
    \end{aligned}
        \label{EL_thermo}
    \end{equation}
The equations \eqref{EL_thermo} are formulated in terms of the quantities $(\bq, \bv, T)$ which are now independent variables, and the non-observable quantities $(\bp, S)$ are connected to these observable variables through the expressions \eqref{G_deriv_pS}.

\begin{remark}
{\rm  The total energy 
\begin{equation} 
E(\bq, \bv, T) = \bp \cdot \bv + S T - G
\label{E_thermo_Lagr}
\end{equation} 
is conserved by equations \eqref{EL_thermo}. 
Indeed, 
\begin{equation}
\begin{aligned}
\dot E &= \dot \bp \cdot \bv + \bp \cdot \dot \bv + S \dot T + T \dot S - \pp{G}{\bq} \cdot \bv - \pp{G}{\bv}\cdot \dot \bv - \pp{G}{T} \dot T \\
&= \left( \frac{d}{dt }\pp{G}{\bv} - \pp{G}{\bq} \right) \cdot \bv - \boldsymbol{F}
^{\rm fr} \cdot \bv =0 .
\label{energy_cons}
\end{aligned}
\end{equation}
The energy quantity \eqref{E_thermo_Lagr} is just the Hamiltonian rewritten in the observable variables $(\bq, \bv, T)$. 
}
\end{remark}

Next, we shall prove that functions $G$ and $\boldsymbol{F}^{\rm fr}$ corresponding to the dynamics \eqref{EL_thermo} are not unique and are defined only up to certain transformations.

\begin{lemma}[On non-uniqueness of solutions]
\label{lem:sol_G_non_unique}
{\rm 
    Suppose a trajectory in the observable space $(\bq(t), \bv(t), T(t))$ is a solution of \eqref{EL_thermo} with $G=G_*(\bq, \bv, T)$, $\boldsymbol{F}^{\rm fr} =\boldsymbol{F}^{\rm fr}_*(\bq, \bv, T)$ with some initial conditions. Then, the same trajectory is a solution for 
    \begin{enumerate} 
        \item (\emph{Scale invariance}) $G=k G_*(\bq, \bv, T)$, $\boldsymbol{F}^{\rm fr} =k \boldsymbol{F}^{\rm fr}_*(\bq, \bv, T)$ for an arbitrary $k \in \mathbb{R}$. 
        \item (\emph{Affine invariance of entropy and Null-Lagrangian}) $G=G_*+S_0 T + \pp{f}{\bq} \cdot \bv$, and the force of friction remain unchanged, where $S_0$ is an arbitrary constant and $f(\bq)$ is arbitrary function of coordinates. 
    \item (\emph{Potential gauge}) $G=G_*+T\varphi(\bq)$, $\boldsymbol{F}^{\rm fr}  = \boldsymbol{F}^{\rm fr}_* -T \pp{\varphi}{\bq}$, where $\varphi(\bq)$ is an arbitrary function of the coordinates.
    \end{enumerate} 
    }
\end{lemma}
\emph{Proof} 
The proof of the scaling and affine invariance parts is a direct consequence of equations \eqref{EL_thermo}. Notice that affine invariance physically corresponds to the freedom in the choice of the origin in entropy, and adding a term to the Lagrangian that is a full time derivative of another function. Thus, the entropy is only defined up to a constant, and momentum is only defined up to a gradient of an arbitrary scalar function $f(\bq)$.

For the third part, substitution of $G=G_*(\bq, \bv, T)+T\varphi(\bq)$, $\boldsymbol{F}^{\rm fr} = \boldsymbol{F}^{\rm fr}_*-T \pp{\varphi}{\bq}$ into the system \eqref{EL_thermo} gives 
\begin{equation}
    \begin{aligned}
        \frac{d}{d t} &  \pp{G_*}{\bv} = \pp{G_*}{\bq}  + T \pp{\varphi}{\bq}  + \boldsymbol{F}^{\rm fr}_* - T \pp{\varphi}{\bq} \, , 
        \\    
        T\frac{d}{d t} &  \left( \pp{G_*}{T} + \varphi(\bq) \right) = - \boldsymbol{F}^{\rm fr}_* - T \pp{\varphi}{\bq}\cdot \bv. 
    \end{aligned}
    \label{augmented_EL_thermo}
\end{equation}
Noticing that $\frac{d}{dt} \varphi(\bq) = \pp{\varphi}{\bq} \cdot \dot \bq$, we see that equations \eqref{augmented_EL_thermo} are the same as the ones satisfied by $G_*$, $\boldsymbol{F}^{\rm fr}_*$. 
$\blacksquare$

\begin{remark}[On non-uniqueness for the case of several temperatures]
\label{remark:invar_several_temp}
    {\rm If the system possesses several subsystems with corresponding temperatures $T_1, \ldots T_N$, and friction forces $\boldsymbol{F}^{\rm fr}_k$, then the corresponding affine and potential gauge invariance is
        \begin{equation}
    G= G_* + \sum_{k=1}^N S_{0,k} T_k + \boldsymbol{P}_0 \cdot \bv +\sum_{k=1}^N T_k \varphi_k(\bq), \quad 
    \boldsymbol{F}^{\rm fr}_k = \boldsymbol{F}^{\rm fr}_{*,k} - T_k\pp{\varphi_k}{\bq}\, .
        \label{inv_several_temp}
    \end{equation}
    This will be the example of two containers connected with an adiabatic piston that we will consider in Sec~\ref{sec:adiabatic-piston}. 
    }
\end{remark}

Therefore, because of the freedom that is inherent to the description in the observable variables, reconstruction of the actual functions $G$ and $\boldsymbol{F}^{\rm fr}$ is challenging and can hardly be accomplished without additional extensive knowledge of the system. However, learning the mapping of the solutions is a tractable problem. 

\medskip

We now turn our attention to deriving a variational integrator based on the variational principle \eqref{Lagr_var_principle_thermo}. As usual, we approximate the time derivatives with discrete derivatives and write \eqref{Lagr_var_principle_thermo} in the discrete formulation as 
\begin{equation}
\begin{aligned}
& \delta \sum  G(\bq_k, \bv_k, T_k) - T_k S_k =0 \quad \mbox{subject to }
\\
& T_k \frac{S_{k+1}-S_k}{h}  = - \boldsymbol{F}^{\rm fr}_k \cdot \bv_k  \, , \quad \bv_k = \frac{\bq_{k+1}-\bq_k}{h} \\ 
&T_k \delta S_k = - \boldsymbol{F}^{\rm fr}_k \cdot \delta \bq_k \, , \quad \delta \bv_k = \frac{\delta \bq_{k+1}-\delta \bq_k}{h} \,,    
\end{aligned}
\label{Lagr_var_principle_discrete_thermo}
\end{equation}  
\rem{ 
\todo{\color{magenta}FGB: It seems that, like in continuous case, $S_k$ is a priori independent and the variations of $T_k$ gives $S_k= \frac{\partial G}{\partial T}(\bq_k, \bv_k, T_k)$ (which is even simpler), so simply:
\begin{equation}
\begin{aligned}
& \delta \sum  G\big(\bq_k, \frac{ \bq_{k+1}- \bq_k}{h}, T_k\big) - T_k S_k =0 \quad \mbox{subject to }
\\
& T_k \frac{S_{k+1}-S_k}{h}  = - \boldsymbol{F}^{\rm fr}_k \cdot \bv_k  \, ,\\ 
&T_k \delta S_k = - \boldsymbol{F}^{\rm fr}_k \cdot \delta \bq_k \,.   
\end{aligned}
\end{equation} 
Also, I insert $\bv_k= \frac{ \bq_{k+1}- \bq_k}{h}$ directly in the function $G$, ok for you or you had something else in mind?
\\ \textcolor{blue}{Yes, I used that assumption too. But in principle $\mathbf{v}_k$ may actually be another variable not equal to $(\bq_{k+1}-\bq_k)/h$. For example, for the data, one could compute $\mathbf{v}_k$ from interpolation of a smooth trajectory. This is the approach I used in simulations. The choice $\bv_k = (\bq_{k+1}-\bq_k)/h$ of course zeros out the first equation, but may not be optimal for the other two (or rather three equations because there are two entropies). So in the variational integrator for the double piston, I solve four equations for $(q, v, T_1, T_2)$ rather than three $(q, T_1, T_2)$.} 
}
}
\rem{ 
\begin{framed}
\textcolor{magenta}{FGB: Above I would like to replace the first condition by 
\[
 \delta \sum  G(\bq_k, \bv_k, T_k) - T_k S_k =0
\]
because $S_k$ is a priori an independent variable. Its variation then gives
\[
T_k\delta S_k
\]
which is needed to use the constraint, and the variation $\delta T_k$ does give its expression $S_k= \frac{\partial G}{\partial T}(\bq_k, \bv_k, T_k)$ \\ 
\textcolor{blue}{OK, good point  - done. In principle, once can have $S_k$ as a dependent variable, then we can use the constraint for $\delta S(\bq_k,\bv_k,T_k)$ as a linear relationship for variations $\de \bq_k$ etc. But the way you are suggesting is more consistent. }}
\end{framed}}
where the subscript $k$, for independent quantities means the $k$-th value of that quantity, and $G_k = G(\bq_k, \bv_k, T_k)$ is the quantity evaluated at $(\bq_k, \bv_k, T_k)$ and similarly for all the derivatives.
The variational principle \eqref{Lagr_var_principle_discrete_thermo} gives: 
\begin{equation}
\begin{aligned}
\frac{1}{h} &  \left[ \bq_{k+1}-\bq_k \right] = \bv_k\\
\frac{1}{h} & \left[ \pp{G}{\bv} (\bq_{k+1}, \bv_{k+1}, T_{k+1}) -\pp{G}{\bv} (\bq_{k}, \bv_{k}, T_{k})\right] = \pp{G}{\bq} (\bq_{k+1}, \bv_{k+1}, T_{k+1})+ \boldsymbol{F}^{\rm fr}_{k+1}\\
\frac{T_k}{h} & \left[ \pp{G}{T}(\bq_{k+1}, \bv_{k+1}, T_{k+1}) - \pp{G}{T} (\bq_{k}, \bv_{k}, T_{k}) \right] = - \boldsymbol{F}^{\rm fr} _k \cdot \bv_k.
\end{aligned}
\label{Lagr_thermal_integrator}
\end{equation}
The case involving multiple entropies and corresponding temperatures can be treated  analogously; the computations are omitted here for brevity. In the next Section, we will approximate $G(\bq, \bv, T)$ and the friction force $\boldsymbol{F}^{\rm fr}$ (or multiple forces in the double-piston case) using a neural network representation. In this setting, equations \eqref{Lagr_thermal_integrator} are used as a learning condition for the neural network function $G_{NN}$ and $\boldsymbol{F}^{\rm fr}_{NN}$.
Before introducing this representation, however, we first highlight a key physical observation regarding the dependence of the force on observable quantities.

\paragraph{On the physics of dissipative forces.}
 In order to complete the problem, we need to approximate the dissipation function $\boldsymbol{F}^{\rm fr}$ in \eqref{Lagr_thermal_integrator}. Our goal is to approximate the friction force using a neural network that can provide an approximation for an arbitrary thermodynamically consistent dissipation function. Remarkably, such a network can be found for a large class of physically relevant friction forces. The most important observation is that the dissipative forces must depend on the observable variables and not on the state variables. This observation will close the system \eqref{Lagr_thermal_integrator}. 
 
Consider, for example, a particle of mass $m$ moving through a fluid or gas according to \eqref{Dissipative_system_general_pq}.  The friction force arises from the fact that air molecules collide with the particle's surface. This force is generated by the mismatch between the \emph{velocities} of air and of the body's surface. Consequently, the friction force $\boldsymbol{F}^{\rm fr}$ can depend only on the \emph{velocities} of the body, not on its momenta $\bp$. Indeed, if the friction force depended on the momenta of the particle, then it would also depend on the the mass of the particle, which is physically implausible. Velocities $\bv$ are obtained by differentiating the Hamiltonian $H$  with respect to momentum: $\bv = \pp{H}{\bp}$.  Similarly, since entropy $S$ is defined only up to an additive constant, the friction force can depend only on temperature $T=\pp{H}{S}$.
Thus, the friction force must be expressed in terms of the observable variables $(\bq, \bv,T)$:
\begin{equation} 
\label{friction_gen} \boldsymbol{F}^{\rm fr} =  \boldsymbol{F}^{\rm fr} (  \bq, \bv,T) \, , \quad \bv := \pp{H}{\bp}, \quad T = \pp{H}{S} \, ,  \quad \mbox{with} \quad 
 \boldsymbol{F}^{\rm fr} \cdot \bv \leq 0 \, ,  
\end{equation} 
and $\boldsymbol{F}^{\rm fr}_k = \boldsymbol{F}^{\rm fr} (\bq_k,\bv_k, T_k)$ in equation \eqref{Lagr_thermal_integrator}. The key challenge is that the observable variables $(\bq, \bv,T)$ differ from the phase space variables $(\bq, \bp,S)$, which is the problem we are addressing in this paper.

\paragraph{Variational integrators on Lie groups using observable data.}
One can readily generalize the approach \eqref{Lagr_var_principle_discrete_thermo}-\eqref{Lagr_thermal_integrator} to thermodynamic systems on Lie groups. We will only do it for the particular case of the motion of a rigid body with friction; the case of a more general Lie group can be constructed analogously. We define 
\begin{equation}
G(\bOm, T) = \bmu(\bOm, T) \cdot \bOm + S(\bOm, T) T - H(\bmu(\bOm,T), S(\bOm,T)) \, . 
    \label{G_SO3}
\end{equation}
Then, defining again $\Omega = \widehat{\bOm}$, we obtain the analogue of \eqref{var_integrator_reduced} for the functions $G$ and $\boldsymbol{f}^{\rm fr}$ expressed in terms of the variables $(\bOm,T)$: 
\begin{equation}
\left\{ 
\begin{aligned}
 & \frac{1}{h} \left( \pp{G}{\bOm_{k+1}} - \pp{G}{\bOm_{k}}\right)  +  \frac{1}{2} \left( \left[\pp{G}{\Omega_{k+1}}, \Omega_{k+1} \right] + \left[ \pp{G}{\Omega_{k}}  , \Omega_{k}\right] \right)^\vee \\ 
 & \qquad \qquad - \frac{h}{4} \left( \Omega_{k+1} \pp{G}{\Omega_{k+1}} \Omega_{k+1} -\Omega_{k} \pp{G}{\Omega_{k}} \Omega_{k} \right)^\vee +    \boldsymbol{f}^{\rm fr}_{k+1} =0 
 \\ 
 &  \frac{1}{h} \left( \pp{G}{T_{k+1}} - \pp{G}{T_k} \right)  =  - \frac{1}{T_k}    \bOm_k \cdot \boldsymbol{f}_k^{\rm fr} \, . 
\end{aligned}
\right. 
    \label{G_var_integrator_SO3}
\end{equation}


\section{Neural networks for the approximation of dissipative forces} 
\label{sec:physics_dissipative_NN}

Our goal for this chapter is to derive a special class of neural networks that can approximate any dissipative force. In other words, we consider the set of all vector-valued functions $\boldsymbol{F}^{\rm rf}(\bq,\bv)$, with the property $\boldsymbol{F}^{\rm fr}\cdot\bv \leq 0$. The considerations developed here can be extended to other vector spaces, multiple friction forces etc. We will make an additional physical assumption that the friction force is regular for small velocities, which is quite physical, since in the limits of slow relative motions, the Stokes' limit of friction forces linear in velocities applies. These dissipative nature of force and regularity at small velocities are the only assumptions that will be necessary to derive these neural network, which will be called the \emph{Dissipative neural networks}. Combining these networks with the power of variational integrators in Section~\ref{sec:var_int} will allow us to develop accurate modeling of dissipative physical systems based exclusively on observable data. 

For convenience, let us write $\boldsymbol{F}^{\rm fr} = - \boldsymbol{\Phi}(\bv, \bq, T)$ and make the following key assumption. 
As we have mentioned above, let us consider forces that, for small values of velocities $\boldsymbol{v}$, do not increase faster than $|\boldsymbol{v}|$. In other words, we consider forces for which the matrix $\mathbb{N}$ defined as 
\begin{equation}
    \label{N_matrix_def} 
    \left|\pp{\Phi_\alpha}{v_\beta}\right| (\bv, \bq, T) < \infty \quad \mbox{as} \quad  |\bv| \rightarrow 0 \, , \quad \Rightarrow \quad  
    \Phi_\alpha (\bv, \bq, T)  = \sum_\beta \mathbb{N}_{\alpha \beta} (\bv, \bq, T) v_\beta   
\end{equation}
where the functions $\mathbb{N}_{\alpha \beta} (\bv, \bq, T)$ are finite as $\bv \rightarrow 0$. Note that this assumption does not mean that $\boldsymbol{\Phi}$ is linear in $\bv$ for $\bv \rightarrow 0$, it just means that the matrix $\mathbb{N}_{\alpha \beta}$ is non-singular as $\bv \rightarrow 0$. For example, both $\boldsymbol{\Phi} = |\bv|^2 \bv$ and $\boldsymbol{\Phi} = \nu \bv$ (Stokes friction) satisfy the requirement \eqref{N_matrix_def}. 
Then, the total dissipation is defined as 
\begin{equation}
    \label{Dissipation_def} D = \bv \cdot \boldsymbol{\Phi}(\bv, \bq, T) = 
    \bv\cdot \mathbb{N}(\bv, \bq, T) \bv = \bv^\alpha \mathbb{N}_{\alpha\beta} (\bv, \bq, T)  \bv^\beta  \geq 0 ,\, \quad \forall \, \, \bv \in \mathbb{R}^n\, . 
\end{equation}
We can split the matrix $\mathbb{N}(\bv, \bq, T)$ into a symmetric and antisymmetric parts  $\mathbb{N} = \mathbb{S} + \mathbb{A}$. Since the antisymmetric matrix  part of the force $\mathbb{A}$ is conservative, it plays no role in the dissipation \eqref{Dissipation_def}, and we assume that $\mathbb{A}$ is already incorporated into the potential energy term of the Hamiltonian. The symmetric part $\mathbb{S}=(\mathbb{N} + \mathbb{N}^T)/2$ of the matrix $\mathbb{N}$ must be a positive definite, symmetric matrix. From linear algebra, we know that 
there exists an orthogonal matrix $\mathbb{Q} $ and a positive definite diagonal matrix $\mathbb{D}$ such that 
\begin{equation}
    \label{Diag_S}  \mathbb{S}  = \mathbb{Q}  \mathbb{D}  \mathbb{Q}^T  \, . 
\end{equation}
We can obtain $\mathbb{Q}(\bv, \bq, T)$ by exponentiating functions $\widehat{q}(\bv, \bq, T)$ taking values in the Lie algebra $\mso(n)$, as $ \mathbb{Q} (\bv, \bq, T)= e^{\widehat{q}(\bv, \bq, T)}$. Elements $\widehat{q} \in \mso(n)$ are $n \times n$ antisymmetric matrices with $n (n-1)/2$ independent components. A natural basis for this space is given by the matrices $\mathbb{E}_{ij}$, which have all entries equal to zero except for $1$ at the $(i,j)$-entry and $-1$ at the $(j,i)$-entry. In this basis, $\widehat{q}$ can be expressed $\widehat{q} (\bv, \bq, T)= \sum_{i<j} q_{ij}(\bv, \bq, T) \mathbb{E}_{ij}$, with no restriction on the values of the function $q_{ij}$.  It is straightforward to verify that any function $\widehat{q}(\bv,\bq,T)$ taking values in $\mso(n)$ can be described that way, and correspondingly, by exponentiating the Lie algebra element, one can describe all functions of  variables $(\bv,\bq,T)$  with the values in $SO(n)$. 

Similarly, we compute the diagonal matrix $\mathbb{D} = {\rm diag}( f_1 ^2, \ldots, f_n ^2) $, for arbitrary functions $f_1 , \ldots, f_n $ with no restrictions on the coefficients. For brevity, we denote $\overline{f} = (f_1, \ldots, f_n)$ and $\mathbb{D} = {\rm diag}\big(\overline{f}^2\big)$. While this representation of diagonal, positive definite matrices is non-unique, it is clear that any function of the form 
$\mathbb{D}(\bv, \bq, T)$ can be expressed in this way. 

Thus, we have proven the following 
\begin{theorem}[Dissipative neural networks]
\label{thm:dissipative_nn}
Consider an $n$-dimensional system \eqref{Dissipative_system_general_pq} with a dissipative function $\boldsymbol{F}^{\rm fr}$. Assume all the conserved forces are contained in the Hamiltonian $H$. Then, all dissipative functions $\boldsymbol{F}^{\rm fr}$ can be approximated by neural networks in three steps: 
\begin{enumerate} 
\item Take a feedforward neural network mapping from $2 n+1$ variables $(\bv, \bq, T)$ with $n(n+1)/2$ outputs. 
\item The first $n (n-1)/2$ outputs describe the antisymmetric matrices 
$\widehat{q} = \widehat{q}_{NN} (\bv, \bq, T) \in \mso(n)$ and $\overline{f} = \overline{f}_{NN}(\bv, \bq, T)$. 
\item The last $n$ outputs represent $\overline{f}_{NN}$ describing the diagonal elements of $\mathbb{D} = {\rm diag}\big( \overline{f}_{NN}^2\big)$ and $\mathbb{Q}= e^{\widehat{q}_{NN}}$. 
\item Compute $\mathbb{S} = \mathbb{Q} \mathbb{D} \mathbb{Q}^T$ and $\boldsymbol{F}^{\rm fr} = - \mathbb{S} \cdot \bv$. 
\end{enumerate} 
\end{theorem} 
In the case when the antisymmetric part of $\mathbb{N}$ in \eqref{Dissipation_def}, which we denote as $\mathbb{A}(\bv, \bq, T)$, does not vanish, we will need to approximate additional unrestricted $n(n-1)/2$ functions $A_{ij}$ to represent $\mathbb{A}(\bv, \bq, T) = \sum_{i<j} A_{ij} (\bv, \bq, T) \mathbb{E}_{ij}$. Here,  as above, $\mathbb{E}_{ij}$ is a matrix that has all zeros except $+1$ at the $(i,j)$ entry and $-1$ at the $(j,i)$ entry. In that case, we will need to approximate $n^2$ functions by the neural network starting with a $2n+1$ dimensional space $(\bv, \bq, T)$. 

\paragraph{Dissipative neural networks for reduced dynamics \eqref{mu_eq_gen}.} If the system allows for a complete reduction as illustrated in \eqref{mu_eq_gen}, Theorem \ref{thm:dissipative_nn} simplifies. In that case, $n$ is now the dimension of reduced momentum $\mu$. The procedure proceeds as follows. 
\begin{enumerate} 
\item Take unrestricted neural network mapping from $n+1$ variables $(\mu, T)$ to $n(n+1)/2$ variables $\widehat{q} = \widehat{q}_{NN} (\mu, T) \in \mso(n)$ and $\overline{f} = \overline{f}_{NN}(\mu, T)$; 
\item Compute $\mathbb{D} = {\rm diag} \big(\overline{f}_{NN}^2\big)$ and $\mathbb{Q}= e^{\widehat{q}_{NN}}$;
\item Take $\mathbb{S} = \mathbb{Q} \mathbb{D} \mathbb{Q}^T$ and $F_\alpha  = - \sum_\beta \mathbb{S}_{\alpha \beta}  \mu^\beta$. 
\end{enumerate}

\begin{remark}[On the Cholesky factorization of matrix S]
    {\rm One could, of course, simply write the matrix $\mathbb{S}$ as a product of two lower-diagonal matrices $\mathbb{L}$ through the well-known Cholesky factorization $\mathbb{S}=\mathbb{L}^T \mathbb{L}$ \cite{higham2009cholesky}. In that case, every element of $\mathbb{L}$ will be described by the output of a neural network. The number of unknown parameters in both cases is exactly the same, and while the formulations are technically equivalent, the Cholesky factorization introduces a hierarchy among the elements of $\mathbb{L}$, which often makes learning difficult. We have found out that the convergence of the algorithm was not satisfactory for the values of parameters we tested when Cholesky factorization was used, whereas the expression $\mathbb{S} = \mathbb{Q} \mathbb{D} \mathbb{Q}^T$ provided a reliable convergence for our test cases. 
    }
\end{remark}

\section{Machine learning through variational integrators}
\label{sec:var_int} 
Now that we have formulated the mathematical background for our method, we are ready to formulate the principles of data-based computing based on the observable quantities. We start with the variational integrator \eqref{Lagr_thermal_integrator} and assume that either $\boldsymbol{F}^{fr}(\bq,\bv,T)$ or $G(\bq,\bv,T)$, or both are unknown. We shall only consider the case of \eqref{Lagr_thermal_integrator}, the variational integrator for the symmetry-reduced case \eqref{G_var_integrator_SO3} is considered in a completely analogous manner. 

One can clearly see that the integrating scheme \eqref{Lagr_thermal_integrator} is implicit, since the information on the step $k$ contains information about the quantities from the step $k+1$. If both $G$ and $\boldsymbol{F}^{fr}$ are known exactly, such scheme necessitates the solution of nonlinear equations which are normally done using Newton or similar methods. In spite of computational complexity compared with explicit methods, the implicit methods are widely used in numerical analysis because of their superior performance in stability \cite{kress2012numerical}. 

Our goal here is different: we assume that the observable data points at the beginning and end of the interval, $(\bq_k,\bv_k,T_k)$ and $(\bq_{k+1},\bv_{k+1},T_{k+1})$, are known, and we use the variational integrator \eqref{Lagr_thermal_integrator} as an equation to determine $G$ and/or $\boldsymbol{F}^{\rm fr}$. We approximate $G=G_{NN}(\bq,\bv,T;\mathbf{W}_G)$ and $\boldsymbol{F}^{\rm fr}=\boldsymbol{F}^{\rm fr}_{NN}(\bq,\bv,T;\mathbf{W}_F)$ as neural networks with parameters $\mathbf{W}_G$ and $\mathbf{W}_F$ respectively, and use equations \eqref{Lagr_thermal_integrator} as conditions to optimize the loss function obtained from equations \eqref{Lagr_thermal_integrator}. Automatic differentiation \cite{baydin2018automatic} is used to compute the appropriate derivatives of $G_{NN}$ in \eqref{Lagr_thermal_integrator}. More precisely, we use the first equation of \eqref{Lagr_thermal_integrator} as the definition of $\bv_k$, and take the loss function $L(\mathbf{W}_G, \mathbf{W}_F)$ as the sum of squares of the second and third equations of \eqref{Lagr_thermal_integrator} (although other loss functions are possible). The parameters of the network are then found by optimization of $L(\mathbf{W}_G, \mathbf{W}_F)$ with respect to the parameters $(\mathbf{W}_G, \mathbf{W}_F)$: 
\begin{equation}
\begin{aligned}
    L(\mathbf{W}_G, \mathbf{W}_F) &=\sum_{k=1}^N \| \mbox{ Second equation of \eqref{Lagr_thermal_integrator}  } \|^2 +\| \mbox{ Third equation of \eqref{Lagr_thermal_integrator}  } \|^2  \, , 
    \\
    (\mathbf{W}_G, \mathbf{W}_F) & = \mbox{arg min } L(\mathbf{W}_G, \mathbf{W}_F) \, . 
    \end{aligned} 
    \label{Loss_var_integrator}
\end{equation}

\paragraph{Advantages of our method.} Since we are writing the relationships  between the beginning and end points as in \eqref{Lagr_thermal_integrator} directly, there is no need to compute the final  point on each interval using some version of ODE solution on that interval as is standard in the previous works on the subject \cite{zhang2022gfinns,gruber2024efficiently} treating the dynamics in the phase space of non-observable variables. In contrast, our loss function \eqref{Loss_var_integrator} provides a direct relationship for computing unknown neural networks for $G_{NN}$ and $\boldsymbol{F}_{NN}$. Of course, once the dynamics in the whole phase space is computed, reconstruction of trajectories in our method will still need solution of implicit equations \eqref{Lagr_thermal_integrator}. 

\paragraph{Non-uniqueness of solutions.} Let us now turn to investigation of uniqueness of solutions $G_{NN}$ and $\boldsymbol{F}^{\rm fr}_{NN}$ when \eqref{Lagr_thermal_integrator} are considered as equations connecting the friction forces and gradients of $G$. Unfortunately, this equation alone is not sufficient to determine both $G$ and $\boldsymbol{F}^{\rm fr}$ uniquely at the same time. Indeed, we can notice a few symmetries of the solution of \eqref{Lagr_thermal_integrator} that are inherited from \eqref{EL_thermo} that prevent the existence of unique solution to these equations. We can formulate the following

\begin{lemma}[On the invariances of solutions]
{\rm 
\label{lem:invariances}
Suppose $G_*$ and $\boldsymbol{F}_*^{\rm fr}$ are solutions of \eqref{Lagr_thermal_integrator}. Then, analogously to Lemma~\ref{lem:sol_G_non_unique}, the following results hold. 
\begin{enumerate}
    \item (\emph{Scale invariance}). For arbitrary $k \in \mathbb{R}$, $k G_*$ and $k \boldsymbol{F}_*^{\rm fr}$ are solutions of \eqref{Lagr_thermal_integrator}. In particular, $G=0$ and $\boldsymbol{F}^{\rm fr} =\mathbf{0}$ are solutions of \eqref{Lagr_thermal_integrator} \emph{for arbitrary data}. 
    \item (\emph{Affine invariance of $G$}) For any solution $G_*$ of \eqref{Lagr_thermal_integrator}, 
    $G = G_* + G_0 + S_0 T + \boldsymbol{P}_0 \cdot \bv$ is also a solution. 
    \item (\emph{Coordinate-temperature shift}) Take any $\mathbf{a} \in \mathbb{R}^3$, and take $G=G_* + T \boldsymbol{a} \cdot \bq$, $\boldsymbol{F}^{\rm fr} = \boldsymbol{F}_*^{\rm fr} - T \boldsymbol{a}$. 
\end{enumerate}
}
\end{lemma}
The proof is obtained by a direct substitution into \eqref{Lagr_var_principle_discrete_thermo}. 

\medskip

Note that the third invariance, adding only a linear term in $\boldsymbol{q}$, is weaker than the corresponding potential gauge invariance of the continuous case, where one can add an arbitrary function of $\boldsymbol{q}$. That reduction in generality is an artifact of the discrete formulation. However, the original full gauge invariance still affects the outcomes of the machine learning procedure.


\begin{remark}
\label{remark:G_F_accuracy_loss}
    {\rm 
    Even though, technically speaking, the third invariance described in the continuous case of Lemma~\ref{lem:sol_G_non_unique} does not exist in the discrete case, the actual invariances of continuous equations may create local minima in the optimization procedure, since they will be 'approximately valid', satisfying the equations in the least square sense. Thus, even though the dynamical solution will be accurate, the solution for $G$ and $\boldsymbol{F}^{\rm fr}$ will have large deviations from their true values. The situation is even more dire in the case of several temperatures, as we have outlined in Remark~\ref{remark:invar_several_temp}. Thus, we believe that searching for the right mapping forward is meaningful, whereas trying to achieve high accuracy of the functions $G$ and $\boldsymbol{F}^{\rm fr}$ using only observable variables is not. We will therefore only discuss the learning of the phase space dynamics and trajectories, but not the actual functions $G$ and $\boldsymbol{F}^{\rm fr}$. 
    }
\end{remark}

\begin{remark}[On the choice of linear term in $\bv$]
\label{linear_term_shift}
    {\rm The choice of a linear term in $G \rightarrow G + \boldsymbol{P}_0 \cdot \boldsymbol{v}$ simply changes the origin of momentum $\bp = \pp{G}{v} \rightarrow p + \boldsymbol{P}_0$. It is common in physics to choose the momentum to vanish whenever velocity vanishes. We do not know that fact \emph{a priori} about any system we consider in the examples in Section~\ref{sec:xamples}; it is a physical assumption that we need to enforce. Thus, we will impose the requirement $\bp=0$ when $\bv=0$ as a soft constraint through taking several 'anchoring points' for $\bv =\mathbf{0}$ for several values of $\bq$ and $T$. We then compute the momentum $\bp = \pp{G}{\bv}$ at these points and add the squared norm $\left\| \pp{G}{\bv}\right\|^2$ to the loss function. The optimization procedure then finds the value of $\boldsymbol{P}_0$ that enforces vanishing momentum for $\bv =\mathbf{0}$.}
\end{remark}

We shall also note that the scaling invariance was noticed in the literature before in the context of metriplectic/GENERIC flows \cite{zhang2022gfinns}. However, as far as we know, the invariances presented here are novel and only pertain to the description of the system in the observable variables case, which is impossible in the metriplectic/GENERIC approach which focuses exclusively on the phase space variables. 
These invariances prevent direct solution of the equations and prevent simultaneous solutions for $G_{NN}$ and obtaining $\boldsymbol{F}^{\rm fr}_{NN}$. Any program trying to simultaneously find $G_{NN}$ and $\boldsymbol{F}^{\rm fr}_{NN}$ will find a solution up to these invariances, leading to arbitrary jumps in the predicted values of values of $\boldsymbol{F}^{\rm fr}$ and the derivatives of $G$, leading to unpredictable values of the momenta $\bp = \pp{G}{\bv}$ and entropies $S = \pp{G}{T}$. The scaling invariance is potentially the most dangerous for machine learning, since it leads, by default, to the zero solution $k=0$.

Thus, it is unrealistic to expect that a full solution of \eqref{Lagr_thermal_integrator} can be found. The observable data does not possess enough information to reconstruct the dynamics. However, if partial information about the system is known, the rest can be obtained from the data and the dynamics in the whole phase space can be computed. 

In what follows, we will assume that either $\boldsymbol{F}^{\rm fr}(\bq,\bv,T)$ or $G(\bq,\bv, T)$ are known exactly and show how our method can learn the dynamics in the phase space and compute the long-term solutions using only a moderate number of points for learning. The partial knowledge we shall utilize will contain experimentally relevant information about thermal properties of the stationary state and physical properties of momentum, such as vanishing momentum for vanishing velocities.

We shall note that we have failed to find a method in the literature capable of learning the evolution of a system based on the observable data, especially using a thermodynamically consistent neural networks like is done here. We thus compare our data with high accuracy results obtained using the Backward Differentiation Formula (BDF), as well as with the variational integrator described in \cite{gay2018variational}.

\section{Examples}
\label{sec:xamples}

In what follows, we present two examples: an adiabatic piston connecting two containers with gas, and the motion of a rigid body with friction. In both cases, we use only observable variables: coordinates (for the first problem), velocities and temperatures. Since there are currently no methods capable of predicting the evolution of thermodynamic systems using these variables as an input for training, we compare the results with the ground truth obtained by a regular numerical integrator.

\subsection{Two containers connected by an adiabatic piston}
\label{sec:adiabatic-piston}

\begin{figure}
    \centering
    \includegraphics[width=0.9\linewidth]{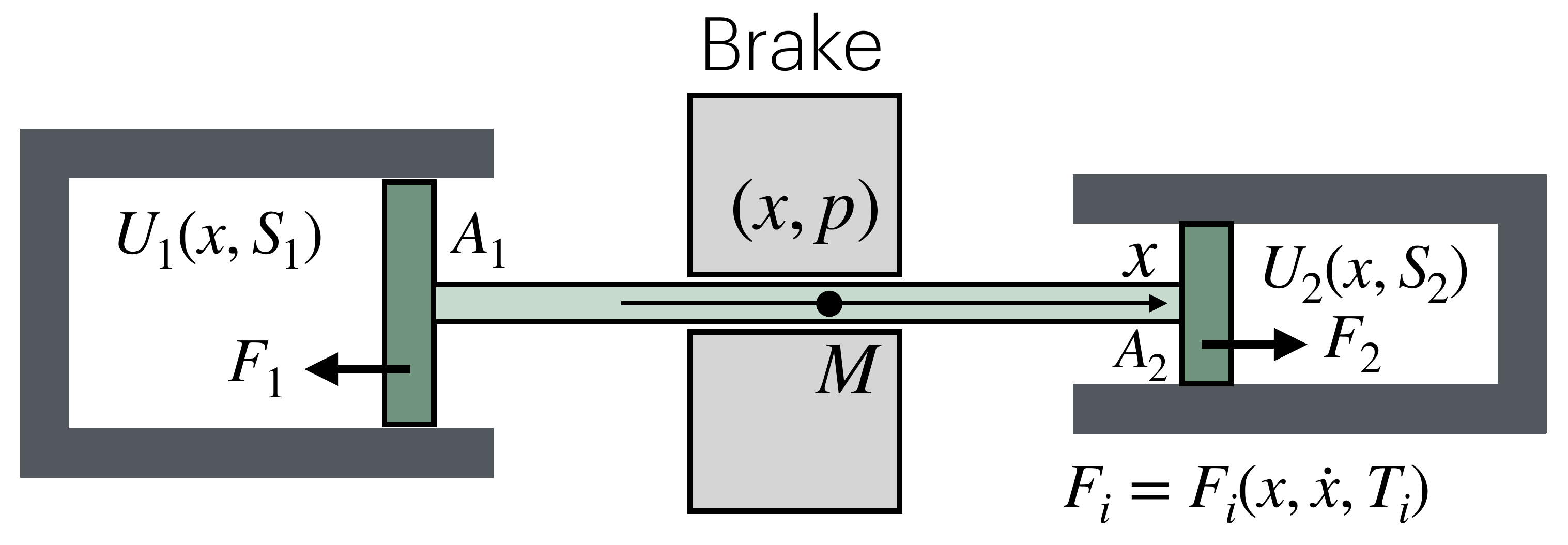}
    \caption{Setup of the problem of an adiabatic piston of mass $M$ separating two chambers. Each chamber $i=1,2$ has the area $A_{1,2}$ and the length $L \pm x$ where $x$ is the coordinate of the piston. Each of the chamber has gas with internal energy depending on the volume $V_i(x)$ and entropy of the gas $S_i$.  }
    \label{fig:Two_chambers_setup}
\end{figure}
Let us consider a system of two containers filled with ideal gas, connected with an adiabatic piston. The sketch of this problem is presented on Figure~\ref{fig:Two_chambers_setup}. When both containers are connected with a piston transferring heat, then the final state is given by the equality of temperatures $T_1 = T_2$. However, when the piston is adiabatic, the answer can get complicated. This problem, initially thought to be simple and even contained in some textbooks on statistical physics, turned out to have quite a bit of unexpected complexity. For the history of the problem and its complete solution see \cite{Gruber1999}. 
Following \cite{Gruber1999}, we write the equations of motion as 
\begin{equation}
    \begin{aligned}
        \dot p & = F^{\rm int}_1 + F^{\rm int}_2 - (\lambda_1+ \lambda_2) v \, , \quad v:= \pp{H}{p}
        \\
        \dot S_1 & = \frac{\lambda_1}{T_1} v^2 \, , \quad 
         \dot S_2  = \frac{\lambda_2}{T_2} v^2 
         \\
         H(x,p,S_1,S_2) & = \frac{1}{2 m} p^2 + \mathsf{U}(S_1,V_1,N_1) + \mathsf{U}(S_2,V_2,N_2)
         \\
         F^{\rm int}_i & = -\pp{\mathsf{U}(S_i,V_i(x),N_i)}{x}
    \end{aligned}
\end{equation}
with the Sackur-Tetrode equation
for the internal energy of ideal gases 
\begin{equation}
    \mathsf{U}(S, V, N) = \left(\widehat{c}\, V\right)^{-2/3} e^{ S/(N k_b)} \, . 
\end{equation}
We normalize $\widehat{c}=1$, although other authors \cite{gruber2024efficiently} take the value $\widehat{c}=102.25$ in the units of inverse volume. As usual,  $k_B$ is the Boltzmann's constant, $N_i$ is the number of molecules in each volume, and the expression for each volume in terms of the position of the piston is written as 
\begin{equation} 
V_1 (x)= A_1 (L+x), \quad  V_2(x) = A_2 (L-x) \, . 
\end{equation} 
We take the parameters 
\begin{equation}
m = 1, \quad A_1 =2, \quad A_2 =1, \quad L=2 \, . 
    \label{params_double_piston}
\end{equation}
With these notations, we obtain 
\begin{equation}
G(x,v,T_1,T_2) = \frac{1}{2} m v^2 + \sum_{i=1,2} N k_B T_i  \log\left[ (\widehat{c} \,V_i)^{2/3} N k_B T_i \right] - N k_B T_i \, , 
    \label{G_double_piston}
\end{equation}
verifying \[ 
p = \pp{G}{v}=mv, \quad 
S_i = \pp{G}{T_i} = N k_B \log\left[ (\widehat{c}\, V_i)^{2/3} N k_B T_i \right] \, . 
\]

We assume that both containers have exactly the same number of molecules, and normalize the units in such a way that $N_i k_B =1$. We also take $A_1=1$ and $A_2=2$. The friction forces provided by the piston on each cylinder are chosen to be:
\begin{equation}
F^{\rm fr}_i = \frac{\lambda_i}{T_i} v\, , \quad 
\lambda_i (x,v,T_i) =\nu_{1,2} +
\kappa_i v^2 \left( 1 + 0.1 T_i^2\right) \, , \quad i=1,2, 
    \label{friction_double_cyl_piston}
\end{equation}
with the parameters $(\nu_1,\nu_2)=(0.02,0.04)$ 
and $(\kappa_1,\kappa_2) = (2,1)$. 

We create three neural networks describing the Hamiltonian $G_{NN}(x,v,T_1,T_2)$ and the forces on each cylinder $F^{\rm fr}_{NN,i}(x,v,T_1,T_2)$, $i=1,2$. The networks for $G$ and each of the friction functions have four inputs, one output. The network for $G$ has four hidden layers of width 128, with the number of trainable parameters being 50,305. The network for $F$ contains three fully connected layers of 128 neurons, with the tanh activation function, and total number of trainable parameters in each network for the force being 33,793. The total number of trainable parameters in all networks for this problem is 117,891. One can notice that because of the choice of our forces, both $\dot S_{1,2} \geq$, which is appropriate because the piston is adiabatic and there is no heat exchange between parts of the system. 

We then generate 100,000 data pairs from 5000 trajectories of length 21 points each, with the initial conditions that are randomly distributed in $(x,p,S_1,S_2)$ with uniform distribution $-1 \leq (x,p) \leq 1$ and $0 \leq (S_1,S_2) \leq 1$. The ground truth data is obtained by performing a high-accuracy BDF simulation from the initial to the final point, with the time between the initial and final point being $h = 0.01$.  
Once we have computed the data pairs, we additionally compute the observable variables $(v,T_1,T_2)$ at the beginning and the end of the interval using the known expression for the Hamiltonian. After computing the variables $(x,v,T_1,T_2)$ at the beginning and the end of each interval, we use only this observable data points, and never use the non-observable data in $(p,S_1,S_2)$. 

We consider three learning scenarios: (a) the function $G(x,v,T_1,T_2)$ is unknown while the friction forces $F_{1,2}^{\rm fr}$ are known; (b) the function $G$ is known while the friction forces  $F_{1,2}^{\rm fr} $ are unknown; and (c) both $G(x,v,T_1,T_2)$ and the forces  $F_{1,2}^{\rm fr}$ are unknown.

Because of the invariances of equations, we need to provide additional conditions on the functions allowing to specify the right properties of the functions $G$ and $F^{\rm fr}$, removing at least some of the discontinuities. We call these measurements 'anchoring points'. We choose the following additional data that can be obtained from experiments: 
\begin{enumerate}
\item For stationary states $v=0$, we assume that one can measure thermal properties at given $q$, $T_1$ and $T_2$. 
    \item For fixed temperatures, $T_1=T_1^0=1$ and $T_2=T_2^0=2$, we assume that the mechanical property (mass) of the system is known. Since the momentum $p = \pp{G}{v}$, then we require that on the specified points in the $(q,T_1,T_2)$ space on the hyperplane $v=0$: 
    \begin{equation}
    \begin{aligned} 
        \mbox{Mass condition:} \, & \pp{p}{v} = \frac{\partial^2 G}{v^2}(v=0) = m \simeq \text{const}
        \\
        \mbox{Momentum consistency:} \, & p (v=0) = \pp{G}{v} (v=0) = 0 .
        \end{aligned} 
    \end{equation}
    \item Information about heat capacity for stationary objects is available: 
    \begin{equation}
        \pp{S}{T} = \frac{\partial^2 G}{\partial T^2} = \frac{C}{T}.
    \label{heat_capacity_cond}
    \end{equation}
\end{enumerate}
Note that the latter condition informs about the second derivative of $G$ with respect to temperature, not imposing any conditions on $S=G_T$ which we consider to be non-measurable. Because of invariances, $S$ is defined only up to a constant.

In all cases, we use Adam algorithm to optmize the parameters of the neural networks. In case a), we use $50000$ training epochs with the learning rate $0.001$, reducing exponentially to $0.0001$ by the end.  The loss function measuring the MSE with anchor conditions, and the residuals of the variational scheme \eqref{Lagr_thermal_integrator} decreases from roughly $\sim 1$ by about 4-5 orders of magnitude, depending on the particular run. 

\begin{figure}
    \centering
    \includegraphics[width=1\linewidth]{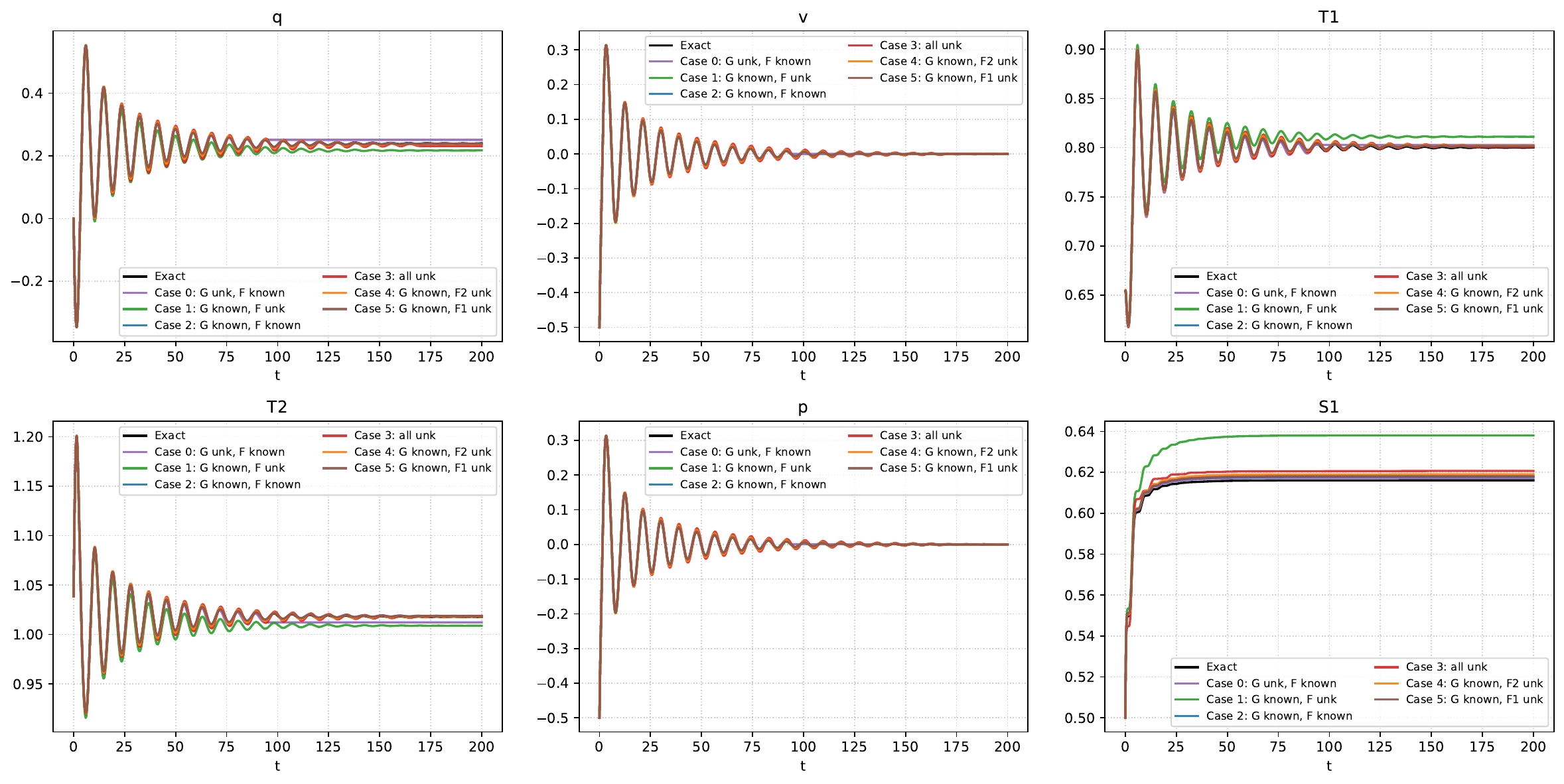}
    \caption{Trajectory reconstruction after learning when partial or minimal information about the system is available: either the information about $G$ or the force. Color notation for solid lines, consistent between both panel of this Figure and Figure~\ref{fig:error_energy_partially_known_GF}. Black: ground truth. Purple: unknown $G$, known $F^{\rm fr}_{1,2}$. Green: known $G$, unknown $F^{\rm fr}_{1,2}$. Blue: both $G$ and $F^{\rm fr}_{1,2}$ are known, representing pure variational integrator. Red: neither $G$ nor $F^{\rm fr}_{1}$ are known. Orange: known $G$ and $F^{\rm fr}_{1}$, unknown $F^{\rm fr}_{2}$. Brown: known $G$ and $F^{\rm fr}_{2}$, unknown $F^{\rm fr}_{1}$. The red line represents the most challenging case, and yet neural network reconstructs the motion with roughly the same accuracy as the pure variational integrator with complete knowledge of the system.  }
    \label{fig:partially_known_GF}
\end{figure}

For validation, on Figure~\ref{fig:partially_known_GF}, we present a trajectory computed from the initial conditions $(x_0,p_0,S_{1,0},S_{2,0})=(0, -0.5, 0.5, 0.5)$, corresponding to $v_0=-0.5$, $T_{1,0} \simeq 0.65$, $T_{1,0} =\simeq .03$. Figure~\ref{fig:partially_known_GF} shows the simulation of 20,000 time steps with the time step $h=0.01$ and plot the resulting trajectory versus the ground truth that was obtained by the same BDF algorithm as the learning data. We remind the reader that the each learning trajectory only consists of 20 data pairs, so the result of the simulations presented are 1000 times longer than the learned trajectory.  We show the observables $(x,v,T_1,T_2)$ and non-observables $(p,S_1)$ (the picture of $S_2$ is similar and not presented here). The blue line represents the ground truth; the black line is purely variational integrator with known $G$ and $F^{\rm fr}_{1,2}$ which is indistinguishable from the blue line. The red line presents the case where the Hamiltonian and hence $G$ is unknown while the friction forces are known. Once the effective approximation of the friction force $F^{\rm fr}_{1,2,NN}$ is found, providing an accurate mapping forward from the data, the variational integrator is used to compute the solution going forward. The green line shows the case when $F^{\rm fr}_{1,2}$ are unknown but the Hamiltonian and hence the function $G$ is known. Here, as well, once the effective approximation of $G_{NN}$ is found, a variational integrator is used to compute the forward solution.

\begin{figure}
    \centering
    \includegraphics[width=1\linewidth]{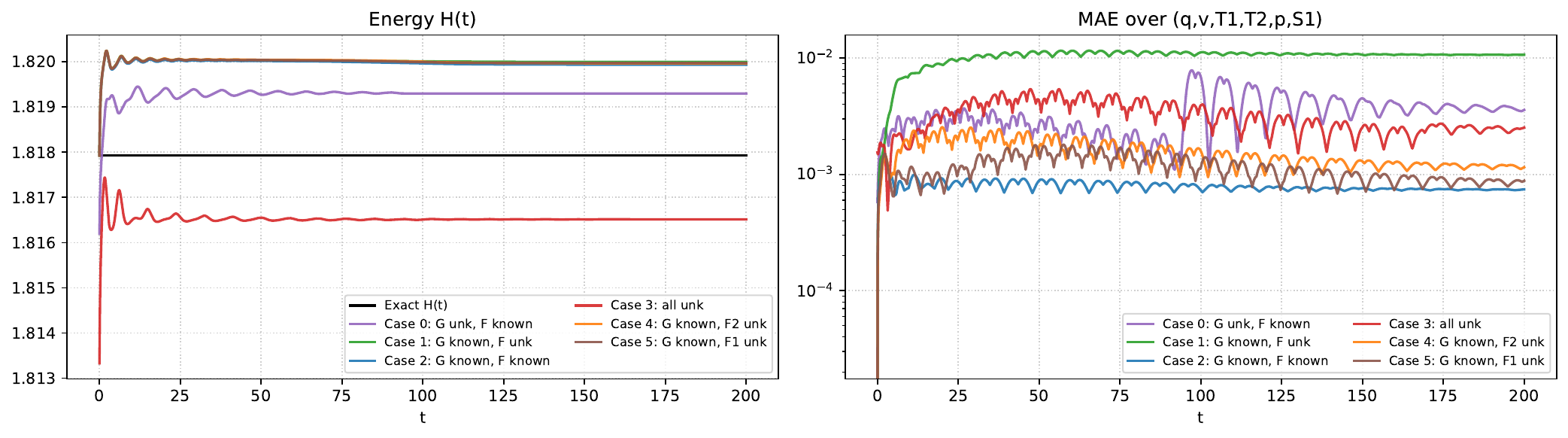}
    \caption{  Errors in energy (left panel) and MAE of individual components(right panel), with all the notations being exactly the same as Figure~\ref{fig:partially_known_GF}. Energy is conserved with high precision and MAE of individual components(right panel) remains bounded. All color notations exactly as in Figure~\ref{fig:partially_known_GF}.
    }
    \label{fig:error_energy_partially_known_GF}
\end{figure}

The graphs for non-observables $p$, $S_1$, and $S_2$ (the latter not shown) are shifted so the final points of the graph correspond to the ground truth, because of the invariance of equations to $G \rightarrow G + p_0 v + S_{1,0} T_1 + S_{2,0} T_2$, which allows to only compute the non-observable quantities $(p,S_1,S_2)$ up to an additive constant.  Note that since we enforce the physicality of momentum (momentum vanishes when velocity vanishes), the ambiguity in momentum definition is removed, but the ambiguity in entropy persists. On  Figure~\ref{fig:error_energy_partially_known_GF}, we present the conservation of energy (left panel) and MAE of the components compared to the exact solution (right panel). One can see from that figure that the errors in energy remain bounded and the errors in components grow slowly, leading to the saturation of errors as on the right panel of Figure~\ref{fig:error_energy_partially_known_GF} . Also, the energy of the system is conserved with a high precision, as shown on the left panel of that Figure.

\color{black} 

\color{black} 

\subsection{A rigid body with friction}

We now turn our attention to the description of the system \eqref{mu_eq_gen} with variational integrator described in \eqref{G_var_integrator_SO3}.
We start with $N = 100000$ data pairs for the beginning and end of short trajectories, $(\bOm_0^i,T_0^i)\rightarrow (\bOm_f^i,T_f^i)$, $i=1, \ldots N$. The time interval between the beginning and end points is taken to be $h =0.1$ for all data. We consider two neural networks, one for approximation of $G_{NN}$ and another for $\boldsymbol{f}_{NN}^{\rm fr}$. Both networks have four inputs for $(\bOm,T)$ and four hidden layers of 64 neurons for $\boldsymbol{f}^{\rm fr}$, and 128 neurons for $G$, each with a tanh activation function. The output for the dissipative neural network to describe $\mathbb{S}(\bv,T)$ as in \eqref{Diag_S} with $\boldsymbol{f}^{\rm fr}_{NN} = - \mathbb{S} \bv$ has six outputs: three variables to describe the infinitesimal rotation and three to describe the diagonals of positive definite matrix. 
The network for $G_{NN}$ has 50,305 parameters, the network for $\boldsymbol{f}^{\rm fr}_{NN}$ has 13,190 parameters, with the total of 63,495 trainable parameters. We denote the parameters for the network describing $G_{NN}$ as $\mathbf{W}_G$, and the parameters describing $\boldsymbol{f}^{\rm fr}$ as $\mathbf{W}_{\boldsymbol{f}}$. 

The loss function is computed as the sum of the norm of the equations \eqref{G_var_integrator_SO3} taken over all the data points: 
\begin{equation}
L(\mathbf{W}_G,\mathbf{W}_{\boldsymbol{f}}) = 
\sum_k \left\| \mbox{ First equation of \eqref{G_var_integrator_SO3} } \right\|^2 + 
\left\| \mbox{ Second equation of \eqref{G_var_integrator_SO3} } \right\|^2 .
    \label{loss_eq_SO3}
\end{equation}

\paragraph{Phase space learning.} 
We start with $N=2000$ data pairs arranged in 100 trajectories of 21 points each. The beginning point of each trajectory is chosen using a uniform random distribution in a cube $| \bmu| \leq 1$ and $0< S<1$. The true Hamiltonian and force function for the system are chosen to be: 
\begin{equation} 
\begin{aligned} 
H(\bmu, S) & = e^{\gamma S} \left( \bmu \cdot \mathbb{I}^{-1} \bmu + U_0 \right)\, , \quad \mathbb{I} = \operatorname{diag}(1,2,3)\, , \quad \gamma = 1 
\\ 
\boldsymbol{f}^{\rm fr}(\bOm, T)&  = -\nu_0 \bOm - \nu_1 \mathbb{A}^T \tanh ( \mathbb{A} \bOm ) \, , \quad 
\mathbb{A} = \left( 
\begin{array}{ccc} 
1& 0.5 & 0.5 \\ 
0.5 & 1 & 0.5 \\ 
0.5 & 0.5 & 1
\end{array} 
\right) \, , \quad \nu_{0,1} =0.01 \, ,
\end{aligned} 
\label{true_H_F}
\end{equation} 
where $\tanh(\mathbf{v})$ just denotes a vector of $(\tanh v_1, \ldots, \tanh v_n)^T$. It is clear that $\boldsymbol{f}^{\rm fr}$ is dissipative since 
\[ 
\boldsymbol{f}^{\rm fr} \cdot \boldsymbol{\Omega} = - \nu_1 |\bOm|^2 - \nu_2  \bxi \cdot \tanh (\bxi) = - \nu_1 |\bOm|^2 - \nu_2 \sum_{i=1}^n  \xi_i \tanh \xi_i \leq 0 \,.
\] 
Incidentally, this friction force comes from the Rayleigh dissipation function $R = \sum_j \ln \cosh \mathbb{A}_{kj} \Omega_{k} $, although this fact plays no role in our calculations below.
For the chosen Hamiltonian \eqref{true_H_F}, the temperature is $T=\pp{H}{S}=\gamma H$. Since $H$ remains constant under the dynamics, $T=\gamma H$ also remains constant. The conservation of temperature will provide another way of demonstrating the accuracy of our results.  
A short calculation yields the following expressions for $G(\bOm,T)$:  
\begin{equation}
    \begin{aligned} 
       \bOm &= e^{\gamma S } \mathbb{I} \bmu \quad \Rightarrow \quad  \bmu = e^{-\gamma S } \mathbb{I} \bOm \\
       \frac{T}{\gamma}&  = e^{-\gamma S } \mathbb{I} \bOm \cdot \bOm + e^{\gamma S} U_0 \\     
        S & = \frac{1}{\gamma} \log \left(\frac{T}{\gamma} + \sqrt{\frac{T^2}{\gamma^2} - 2 \mathbb{I} \bOm \cdot \bOm} \right)
        \\
        G & =e^{-\gamma S(\bOm, T) } \mathbb{I} \bOm \cdot \bOm + T S(\bOm,T) - \frac{T}{\gamma} .
    \end{aligned}
    \label{G_func_SO3}
\end{equation}
We have chosen the root $+$ in the expression for $S$ from the solution of quadratic equation for $e^{\gamma S}$, since the $-$ root gives unphysical result $e^{\gamma S}=0$ for $\bOm =\mathbf{0}$.

Clearly, the function $G$ defined by \eqref{G_func_SO3} is not quadratic in velocities $\bOm$ since $S(\bOm,T)$ depends on velocities in a highly non-linear way. Thus, unlike classical mechanical Lagrangians, the thermodynamic Lagrangians $G(\bOm,T)$ derived here do not have to be quadratic in velocities $\bOm$. This observation shows the power of machine learning, as data-based methods are capable of finding arbitrary dependencies of $G$ on velocities and temperatures from data. 

Three cases are considered: (a) known $G$ and unknown $\boldsymbol{f}^{\rm fr}$; (b) unknown $G$ and unknown $\boldsymbol{f}^{\rm fr}$; and (c) both 
$G$ and  $\boldsymbol{f}^{\rm fr}$ are unknown. In both cases, Adam minimizer is used to compute $G_{NN}$ or $\boldsymbol{f}^{\rm fr}_{NN}$. The  learning rate for both cases is taken to be initially $0.001$, exponentially decreasing to $0.00001$, with $10000$ epochs, reducing the loss function by about four orders of magnitude.
 
\begin{figure}
    \centering
    \includegraphics[width=0.95\linewidth]{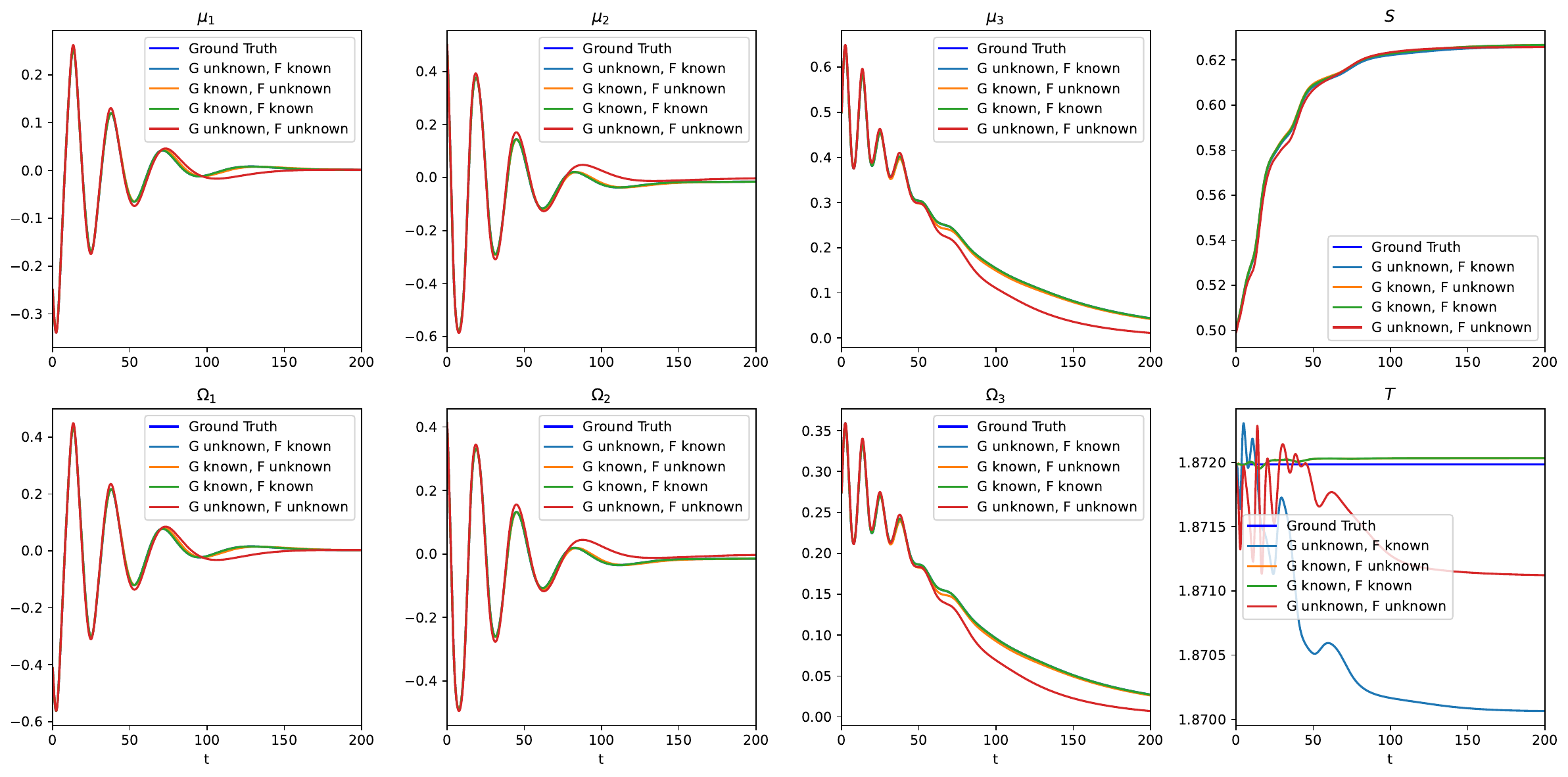}
    \caption{Results of the learning scheme with various known and unknown quantities. Red line: unknown $G$, known $\boldsymbol{f}^{\rm fr}$. Green line: known $G$, unknown $\boldsymbol{f}^{\rm fr}$. Black line: both $\boldsymbol{f}^{\rm fr}$ and $G$ are known (variational integrator) which is the best outcome based on the discrete observable data. Orange line: the most challenging case when both $G$ and $\mathbf{f}^{\rm fr}$ are unknown. Blue line: ground truth. Top panel: non-observable variables: $(\mu_1, \mu_2, \mu_3, S)$. Bottom panel: observables $(\Omega_1, \Omega_2, \Omega_3, T)$. In this particular case, the temperature $T$ is proportional to the energy, so the conservation of temperature reflects the conservation of energy as illustrated on the left panel of Figure~\ref{fig:ThermoNets_Hamiltonian_learning_error}.}
    \label{fig:ThermoNets_Hamiltonian_learning_results}
\end{figure}

\begin{figure}
    \centering
        \includegraphics[width=0.95\linewidth]{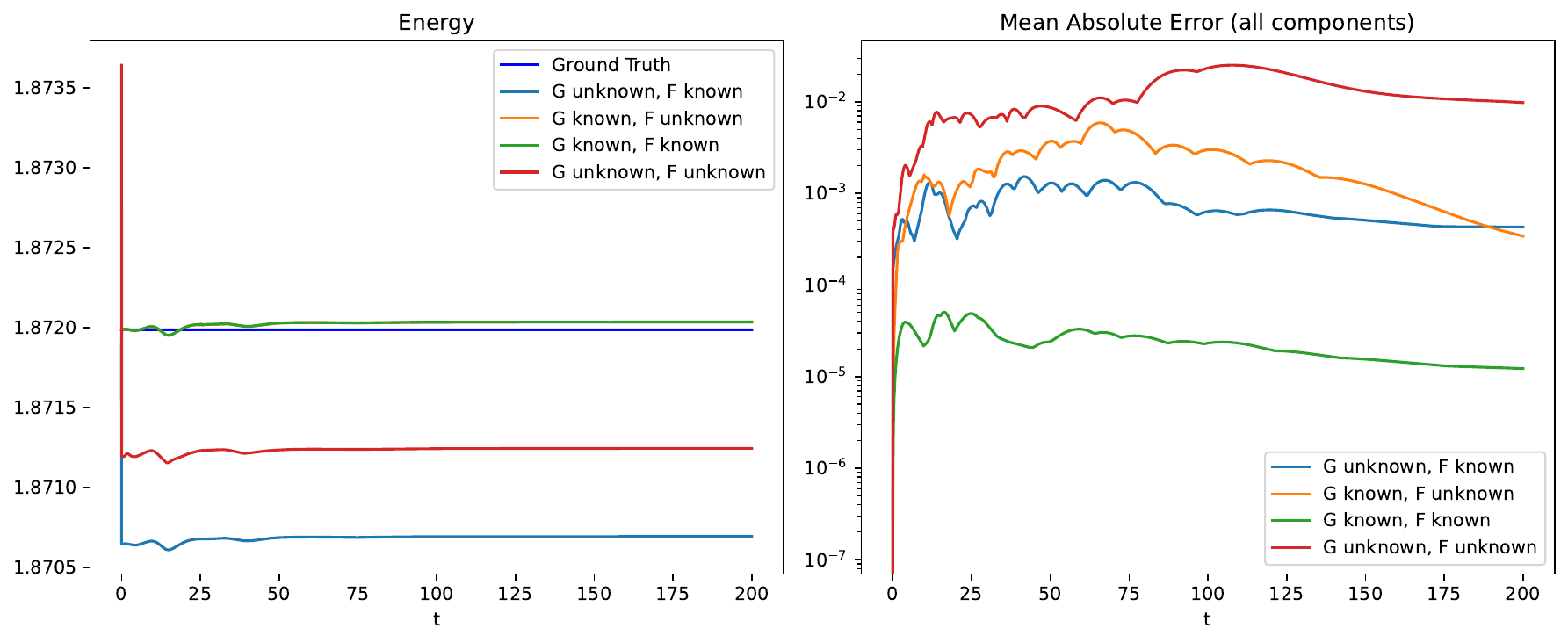}
    \caption{Left panel: energy conservation, taken over 20,000 time steps. Blue line:  ground truth, with the energy measured as a true Hamiltonian. Solid red, green and blue lines correspond to the notation on Figure~\ref{fig:ThermoNets_Hamiltonian_learning_results}. Right panel: MAE taken over all components. }
    \label{fig:ThermoNets_Hamiltonian_learning_error}. 
\end{figure}

For a sample validation trajectory shown on Figure~\ref{fig:ThermoNets_Hamiltonian_learning_results}, choose $\bmu_0 = (-0.25, 0.5, 0.5, 0.5)$ and $S_0=0.5$, corresponding to $\bOm = \mathbb{I}^{-1} \bmu_0 = (-0.41,0.41,0.27)$ and $T_0 \simeq 1.87$. The errors in the Hamiltonian and MAE of observable components $(\bOm,T)$ are presented on the left and the right panels of Figure~\ref{fig:ThermoNets_Hamiltonian_learning_error}, respectively. One can see that the model faithfully reproduces the long-term behavior of the system. 

In order to remove the ambiguity of the scaling and shift gauge of $G$, we introduce a grid of $20\times20\times20$ points in $\bOm$ space, taken for a chosen $T_0 = 2$, and a line of 20 data points data in $\bOm=0$ subspace for $\bOm=0$. We measure the following data in these specified points: the values of $G$, the values of $\bmu = \pp{G}{\bOm}$ and $S=\pp{G}{T}$, and also the value of $\pp{S}{T}=\frac{\partial^2 G}{\partial T^2}$. The physical justification for inclusion of these quantities in the data is as follows. When $T=$const, we can measure simple mechanical properties of the system, \emph{i.e.}, dependence of the purely mechanical Lagrangian on $\bOm$. The inclusion of the value of $S=\pp{G}{T}$ at given temperature for a non-moving object is simply specifying the the origin of $S$, removing the ambiguity of $G \rightarrow G + S_0 T$. Measurement of the specific heat $\pp{S}{T}=\frac{\partial^2 G}{\partial T^2}$ for the non-moving object can also be done in a lab. Our network predicts the value of $G$ in the whole $(\bOm,T)$ space, using the measurable data as anchors to remove the freedom in choosing $G$ outlined in Lemma~\ref{lem:invariances}. 

We used 5000 short trajectories of 21 points each, totaling 100,000 data pairs. The initial points of these trajectories are randomly distributed, with the uniform distribution, in the $|\bmu| \leq 1$ and $0<S<1$ space. The original data are first produced in the $(\bmu, S)$ space and then transferred to the observable space $(\bOm, T)$ using the exact expression for the Hamiltonian \eqref{true_H_F}. 
\\
The network for $F$ contained a neural network with 4 inputs, 4 hidden layers of 64 neurons, and 3 output layers, with 17,190 parameters. The network for $G$ contained 4 inputs, 4 hidden layers of 128 neurons and one output for the scalar function of $G$, with 12,865 parameters. 
\\ 
Even with such a relatively small network, and a modest amount of data, we successfully learn the motion in the phase space. This example shows that the V-NOTS approach shows promise for simulations of long-term dynamics of systems with friction with high accuracy.


\section{Conclusions}

We have derived a novel method for learning dissipative systems from observable data, based on:\\
a) a description of the system based on the thermal Lagrangian and its neural approximation;\\
b) a thermodynamically consistent neural network approximation of dissipative forces; and\\
c) a variational discretization framework as the foundation for constructing accurate mappings in both phase space and the space of observables.

Our approach applies to a broad class of finite-dimensional dissipative systems which may or may not be of the metriplectic form. A key advantage of our method is that the learning process relies solely on observable data, \emph{i.e.}, coordinates, velocities, and temperatures. In contrast, most previous works on thermodynamics systems has focused on phase space variables such as coordinates, momenta, and entropy, with the latter two being generally unobservable in experiments.  This makes our approach particularly well suited for data-driven modeling directly from experimental observations, as opposed to methods that depend purely on phase space variables.

In the future, it would be interesting to extend our method to the discretization of models in continuum mechanics, such as solid mechanics, fluid mechanics, and oceanography. A key challenge in this approach lies in the nature of the available data, which are often measured in terms of \emph{Eulerian} quantities, such as velocities and temperatures at a given location as given, for example, by weather stations. A more promising avenue involves the use of Lagrangian data, such as measurements from drifting ocean buoys, which track the velocity and temperature of (approximately) the same material points in the fluid. Using our method on the Eulerian data faces the difficulty in constructing thermodynamically consistent, structure-preserving discretization in the Eulerian formulation based on the observable quantities only. To address this, we intend to develop a variational formulation, based on the thermal Lagrangian method presented here, in the Lagrangian and Eulerian approach. Extensions of existing variational discretization methods for fluid-based systems \cite{eldred2017total,gay2019variational,ElGB2020,gawlik2024variational} and variational approaches to fluid-structure interactions, such as fluid-conveying elastic tubes \cite{gay2015flexible,gay2016variational,gay2019geometric}, and porous media \cite{farkhutdinov2020geometric,gay2020variational,farkhutdinov2021actively,gay2022variational,gay2024thermodynamically}, present promising directions for integrating both Eulerian and Lagrangian data in future applications of our method.

\section*{Acknowledgements}

We are grateful to Anthony Bloch, Pavel Bochev, Eric Cyr, David Martin de Diego, Tristan Griffith, Darryl D. Holm and Dmitry Zenkov for fruitful and engaging discussions. The work of FGB was partially supported by a startup grant from Nanyang Technological University and by the Ministry of Education, Singapore, under Academic Research Fund (AcRF) Tier 1 Grant RG99/24. The work of VP was partially supported by the NSERC Discovery Grant 2023-03590, and the University of Alabama Shelby Funds. The authors acknowledge the use of the High Performance Computing (HPC) resources and services provided by University of Alabama Research Computing.

This article has been co-authored by an employee of National Technology \& Engineering Solutions of Sandia, LLC under Contract No. DE-NA0003525 with the U.S. Department of Energy (DOE). The employee owns right, title, and interest in and to the article and is responsible for its contents. The United States Government retains and the publisher, by accepting the article for publication, acknowledges that the United States Government retains a non-exclusive, paid-up, irrevocable, world-wide license to publish or reproduce the published form of this article or allow others to do so, for United States Government purposes. The DOE will provide public access to these results of federally sponsored research in accordance with the DOE Public Access Plan https://www.energy.gov/downloads/doe-public-access-plan.

\section*{Competing interests and availability of codes and data}
The authors declare no competing interests. 
\\
All the codes used in this paper are available by request to the corresponding author. No external data sets were used; all data sets were generated by the codes as described. 

\section*{Declaration of Generative AI Use}
All theoretical computations were performed by the authors. The code for the examples in Section~\ref{sec:xamples} was initially developed manually and then optimized using \emph{Gemini Pro}. The resulting code improvements were thoroughly verified to ensure accuracy and reproducibility.

\bibliographystyle{unsrt}
\bibliography{References_LPNets}

\appendix

\section{The case of multiple entropies}
\label{app:mutl_entropies}
\paragraph{Thermodynamic systems with several entropies.} Let us consider a Lagrangian with dependencies on mechanical variables as well as several entropies, i.e., $L(\bq, \dot \bq, S_1,...,S_P)$. We assume that the system is subject to friction forces $\boldsymbol{F}^{\rm fr (i)}(\bq, \dot{\bq},S_1,..., S_P)$ and entropy fluxes $J_{ij}=J_{ji}$. The above approach is extended as follows:
\begin{equation}\label{VCond_ns}
\delta \int_0^T L( \bq, \dot{\bq},S_1,...,S_P) + \sum_{i=1}^P \dot \Gamma^i(S_i-\Sigma_i) {\rm d}t=0
\end{equation}
subject to the constraints
\begin{equation}\label{PC_ns}
\frac{\partial L}{\partial S_i}\dot \Sigma_i = \boldsymbol{F}^{\rm fr (i)}\cdot \dot\bq+\sum_jJ_{ij}\dot\Gamma^j , \quad i=1,...,P
\end{equation}
and for variations subject to
\begin{equation}\label{VC_ns}
\frac{\partial L}{\partial S_i}\delta \Sigma_i = \boldsymbol{F}^{\rm fr (i)}\cdot \delta\bq+\sum_jJ_{ij}\delta\Gamma^j , \quad i=1,...,P
\end{equation}
with $\delta\bq(0)=\delta\bq(T)=0$. This principle yields the equations
\[
\frac{d}{dt}\frac{\partial L}{\partial\dot{\bq}}- \frac{\partial L}{\partial\bq}= \sum_{i=1}^P\boldsymbol{F}^{\rm fr (i)}, \quad \frac{\partial L}{\partial S_i}\dot S_i = \sum_{j=1}^P\boldsymbol{F}^{\rm fr(i) }\cdot \dot\bq +\sum_{j=1}^P J_{ij}\left(\frac{\partial L}{\partial S_i}-\frac{\partial L}{\partial S_j} \right),
\]
see \cite{GBYo2023}. This setting will be applied to the adiabatic piston system in Section \ref{sec:adiabatic-piston}, where the meaning of the variables $\Gamma^j$ and $\Sigma_i$ will be clarified.

\end{document}